\documentclass[letterpaper]{article}
\usepackage{aaai}
\usepackage{times}
\usepackage{helvet}
\usepackage{courier}
\usepackage{wrapfig}
\usepackage{cleveref}

\usepackage{amsmath}
\usepackage{amssymb}
\usepackage{algorithm}
\usepackage{algpseudocode}
\usepackage{tikz}

\newtheorem{theorem}{Theorem}
\newtheorem{definition}{Definition}
\def\myproof{1} 
\begin{document}
\crefformat{footnote}{#2\footnotemark[#1]#3}
\newcommand{\set}[1]{{\mathcal #1}}
\newcommand*\circled[1]{%
  \tikz[baseline=(C.base)]\node[draw,circle,inner sep=0.1pt](C) {\tiny$#1$};\!
}
\title{GP-Localize: Persistent Mobile Robot Localization using\\Online Sparse Gaussian Process Observation Model}

\author{
Nuo Xu$^{\dag}$ \and Kian Hsiang Low$^{\dag}$ \and Jie Chen$^{\S}$ \and Keng Kiat Lim$^{\dag}$ \and Etkin Bari\c{s} \"{O}zg\"{u}l$^{\dag}$\\
Department of Computer Science, National University of Singapore, Republic of Singapore$^{\dag}$\\
Singapore-MIT Alliance for Research and Technology, Republic of Singapore$^{\S}$\\
\{xunuo, lowkh, kengkiat, ebozgul\}@comp.nus.edu.sg$^{\dag}$, chenjie@smart.mit.edu$^{\S}$
}
\maketitle

\begin{abstract}
\vspace{-2mm}
\begin{quote}
Central to robot exploration and mapping is the task of persistent localization in environmental fields characterized by spatially correlated measurements.
This paper presents a \emph{Gaussian process localization} (GP-Localize) algorithm that, in contrast to existing works, can exploit the spatially correlated field measurements taken during a robot's exploration (instead of relying on prior training data) for efficiently and scalably learning the GP observation model online through our proposed novel online sparse GP.
As a result, GP-Localize is capable of achieving \emph{constant} time and memory (i.e., independent of the size of the data) per filtering step, which demonstrates the practical feasibility of using GPs for persistent robot localization and autonomy.
Empirical evaluation via simulated experiments with real-world datasets and a real robot experiment shows that GP-Localize outperforms existing GP localization algorithms.\vspace{-2mm}
\end{quote}
\end{abstract}

\section{Introduction}
\label{sect:intro}
\vspace{-0.5mm}
Recent research in robot exploration and mapping has focused on developing adaptive sampling and active sensing algorithms \cite{LowAAMAS13,LowRSS13,LowUAI12,LowICML14,LowAAMAS08,LowICAPS09,LowAAMAS11,LowICRA07,LowAAMAS12,LowAAMAS14} to gather the most informative data/observations for modeling and predicting spatially varying environmental fields that are characterized by \emph{continuous-valued}, \emph{spatially correlated} measurements. 
Application domains (e.g., environmental sensing and monitoring) requiring such algorithms
often contain multiple fields of interest: (a) Autonomous underwater and surface vehicles are tasked to sample ocean and freshwater phenomena including temperature, salinity, and oxygen concentration fields 
\cite{LowSPIE09,LowAeroconf10},
(b) indoor environments are spanned by temperature, light, and carbon dioxide concentration fields that affect the occupants' comfort and satisfaction towards the environmental quality across different areas,
and (c) WiFi access points/hotspots situated at neighboring locations produce different but overlapping wireless signal strength fields over the same environment. 
These algorithms operate with an assumption that the locations of every robot and its gathered observations are known and provided by its onboard sensors such as the widely-used GPS device.
However, GPS signals may be noisy (e.g., due to urban canyon effect between tall buildings) or unavailable (e.g., in underwater or indoor environments).
So, it is desirable to alternatively consider exploiting the spatially correlated measurements taken by each robot for localizing itself within the environmental fields during its exploration; this will significantly extend the range of environments and application domains in which a robot can localize itself.

To achieve this, our robotics community will usually make use of a probabilistic state estimation framework known as the \emph{Bayes filter}: 
It repeatedly updates the belief of a robot's location/state by assimilating the field measurements taken during the robot's exploration through its \emph{observation model}.
To preserve time efficiency,
the Bayes filter imposes a Markov property on the observation model:
Given the robot's current location, its current measurement is conditionally independent of the past measurements. 
Such a Markov property is severely violated by the spatial correlation structure of the environmental fields, thus strongly degrading the robot's localization performance.
To resolve this issue, the works of \citeauthor{Ko2008} \shortcite{Ko2008,Ko2009} have 
integrated a rich class of Bayesian nonparametric models called the \emph{Gaussian process} (GP) into the Bayes filter, which allows the spatial correlation structure between measurements to be formally characterized (i.e., by modeling each field as a GP) and the observation model to be represented by fully probabilistic predictive distributions (i.e., one per field/GP) with formal measures of the uncertainty of the predictions. 

Unfortunately, such expressive power of a GP comes at a high computational cost, which hinders its practical use in the Bayes filter for persistent robot localization: It incurs cubic time and quadratic memory in the size of the data/observations.
Existing works  \cite{Brooks2008,Ferris2006,Ferris2007,Ko2008,Ko2009}
 have sidestepped this computational difficulty by assuming the availability of data/observations \emph{prior} to exploration and localization for training the GP observation model offline; some \cite{Brooks2008,Ferris2006,Ko2008} 
 have assumed these given prior measurements to be labeled with known locations while others \cite{Ferris2007,Ko2009} have inferred their location labels.
The Markov assumption on the observation model can then be ``relaxed'' to 
conditional independence between the robot's current measurement and past measurements (i.e., taken during its exploration) given its current location and the trained GPs using prior data/observations, thus improving the efficiency at each filtering step during its exploration to quadratic time in the size of the prior training data.
Any measurement taken during the robot's actual exploration and localization is thus not used to train the GP observation model.
Such a ``relaxed'' Markov assumption may hold in certain static environments. However, it becomes highly restrictive and  is easily violated in general, practical environmental settings  
where, for example, 
(a) limited sampling budget (i.e., in terms of energy consumption, mission time, etc.) 
forbids the collection of prior training data or only permits extremely sparse prior data to be gathered relative to a large environment, thus resulting in an inaccurately trained GP observation model,
(b) environmental changes invalidate the prior training data, and (c) the robot's actual exploration path is spatially distant from the prior observations, hence making the trained GP observation model uninformative to its localization.
All these practical considerations motivate us to tackle a fundamental research question: 
Without prior training data, how can GPs be restructured to be used by a Bayes filter for \emph{persistent} robot localization in environmental fields characterized by spatially correlated measurements?

This paper presents a \emph{Gaussian process localization} (GP-Localize) algorithm that, in contrast to existing works mentioned above, can exploit the spatially correlated field measurements taken during a robot's exploration (instead of relying on prior training data) for efficiently and scalably learning the GP observation model online through our proposed novel online sparse GP 
(Section~\ref{sect:gplocalize}).
As a result, GP-Localize is capable of achieving \emph{constant} time and memory (i.e., independent of the size of the data/observations) per filtering step, which we believe is an important first step towards demonstrating the practical feasibility of employing GPs for persistent robot localization and autonomy.
We empirically demonstrate through simulated experiments with three real-world datasets as well as a real robot experiment that GP-Localize outperforms existing GP localization algorithms (Section~\ref{sect:expt}).
\vspace{-3.5mm}
%
\section{Background}
\label{sect:bkgd}
\vspace{-0.5mm}
\subsection{Modeling Environmental Field with GP}
\label{sect:gp}
\vspace{-0.5mm}
The Gaussian process (GP) can be used to model an environmental field as follows\footnote{To simplify exposition, we only describe the GP for a single field; for multiple fields, we assume independence between them to ease computations.}: The environmental field is defined to vary as a realization of a GP. Let $\set{X}$ be a set of locations representing the domain of the environmental field such that each location $x\in\set{X}$ is associated with a realized (random) field measurement $z_x (Z_x)$ if $x$ is observed (unobserved). Let $\lbrace Z_x\rbrace_{x\in\set{X}}$ denote a GP, that is, every finite subset of $\lbrace Z_x\rbrace_{x\in\set{X}}$ has a multivariate Gaussian distribution \cite{Rasmussen2006}. The GP is fully specified by its \emph{prior} mean $\mu_x \triangleq \mathbb{E}[Z_x]$ and covariance $\sigma_{xx'} \triangleq$ cov[$Z_x, Z_{x'}$] for all $x, x' \in \set{X}$,
the latter of which characterizes the spatial correlation structure of the field and can be defined using a covariance function.
A common choice is the squared exponential covariance function
 $
 \sigma_{xx'} \triangleq \sigma^2_s \exp\{-0.5 (x - x')^{\top} M^{-2} (x - x')+\sigma_n^2\delta_{xx'}\} 
$
where $\sigma_s^2$ and $\sigma_n^2$ are, respectively, the signal and noise variance controlling the intensity and the noise of the measurements, $M$ is a diagonal matrix with length-scale components $\ell_1$ and $\ell_2$ controlling, respectively, the degree of spatial correlation or ``similarity'' between measurements in the horizontal and vertical directions of the field, and 
$\delta_{xx'}$ is a Kronecker delta of value $1$ if $x = x'$, and 0 otherwise. 

A chief advantage of using the full GP to model the environmental field is its capability of performing probabilistic regression: 
Supposing a robot has visited and observed a set $\set{D}$ of locations and taken a column vector $z_{\set{D}}$ of  corresponding realized measurements, the full GP can exploit these observations to predict the measurement at any unobserved location $x\in\mathcal{X}\setminus\mathcal{D}$ as well as provide its corresponding predictive uncertainty using a Gaussian predictive distribution $p(z_x|x,\set{D},z_\set{D}) = \mathcal{N}(\mu_{x|\set{D}}, \sigma_{xx|\set{D}})$ with the following \emph{posterior} mean and variance, respectively:
\vspace{-1mm}
\begin{equation}
\displaystyle\mu_{x|\set{D}}\triangleq \mu_x + \Sigma_{x\set{D}}\Sigma_{\set{D}\set{D}}^{-1}\left( z_\set{D} - \mu_\set{D} \right)
\label{mean}
\vspace{-2mm}
\end{equation}
\begin{equation}
\displaystyle\sigma_{xx|\set{D}}\triangleq\sigma_{xx}-\Sigma_{x\set{D}}\Sigma_{\set{D}\set{D}}^{-1}\Sigma_{\set{D}x}
\label{variance}
\end{equation}
where 
${\mu}_\set{D}$ is a column vector with mean components $\mu_{x'}$ for all $x'\in \set{D}$,
$\Sigma_{x\set{D}}$ is a row vector with covariance components $\sigma_{xx'}$ for all $x'\in \set{D}$, $\Sigma_{\set{D}x}$ is the transpose of $\Sigma_{x\set{D}}$, and $\Sigma_{\set{D}\set{D}}$ is a matrix with components $\sigma_{x'x''}$ for all $x',x'' \in \set{D}$. 
\vspace{-1mm}
%
\subsection{Sparse Gaussian Process Approximation}
\label{sect:sgp}
%
The key limitation hindering the practical use of the full GP in the Bayes filter 
for persistent robot localization is its poor scalability in the size $|\set{D}|$ of the data/observations: 
Computing the Gaussian predictive distribution (i.e., \eqref{mean} and \eqref{variance}) requires inverting the covariance matrix $\Sigma_{\set{D}\set{D}}$, which incurs $\set{O}(|\set{D}|^3)$ time and $\set{O}(|\set{D}|^2)$ memory.
To improve its scalability, GP approximation methods \cite{LowUAI12,LowUAI13,joaquin2005} have been proposed, two of which will be described below.

The simple
sparse \emph{subset of data} (SoD) approximation method uses only a subset $\set{S}$ of the 
set $\set{D}$ of locations (i.e., $\set{S}\subset \set{D}$) observed and the realized measurements $z_\set{S}$ taken by the robot
to produce a Gaussian predictive distribution of the measurement at any unobserved location $x\in \set{X}\setminus\set{D}$ with the following posterior mean and variance, which are similar to that of full GP
(i.e., by replacing $\set{D}$ in (\ref{mean}) and (\ref{variance}) with $\set{S}$):
\vspace{-2mm}
%
\begin{equation}
  \mu_{x|\set{S}}=\mu_x+\Sigma_{x\set{S}}\Sigma_{\set{S}\set{S}}^{-1} (z_\set{S} - \mu_\set{S}) 
\label{eq:fsod}
\vspace{-0.7mm}
\end{equation}
\begin{equation}
  \sigma_{xx|\set{S}}=\sigma_{xx}-\Sigma_{x\set{S}}\Sigma_{\set{S}\set{S}}^{-1}\Sigma_{\set{S}x}\ .
\label{pcovsod}
\end{equation}
The covariance matrix $\Sigma_{\set{S}\set{S}}$ is inverted using $\set{O}(|\set{S}|^3)$ time and $\set{O}(|\set{S}|^2)$ memory, which are independent of $|\set{D}|$.  
The main criticism of SoD is that it does not exploit all the data for computing the Gaussian predictive distribution, thus yielding an unrealistic overestimate \eqref{pcovsod} of the predictive uncertainty (even with fairly redundant data and informative subset $\set{S}$) \cite{joaquin2005} and in turn an inaccurately trained observation model.

The sparse \emph{partially independent training conditional} (PITC) approximation method is the most general form of a class of reduced-rank covariance matrix approximation methods reported in \cite{joaquin2005} exploiting the notion of a support set $\set{S}\subset\set{X}$.
Unlike SoD, PITC can utilize all data (i.e., $\set{D}$ and $z_\set{D}$) to derive a Gaussian predictive distribution of the measurement at any $x\in \set{X}\setminus\set{D}$ with the following posterior mean and variance:\vspace{-0.5mm}
\begin{equation}
\mu^{\text{PITC}}_{x| \set{D}} \triangleq \mu_x + \Gamma_{x\set{D}}(\Gamma_{\set{D}\set{D}} + \Lambda)^{-1}(z_\set{D}-\mu_\set{D}) 
\label{PITCmean}
\end{equation}
\begin{equation}
\sigma^{\text{PITC}}_{xx| \set{D}} \triangleq \sigma_{xx} - \Gamma_{x\set{D}}(\Gamma_{\set{D}\set{D}} + \Lambda)^{-1}\Gamma_{\set{D}x} 
\label{PITCvar}
\end{equation}
where $\Gamma_{\set{A}\set{A}'}=\Sigma_{\set{A}\set{S}}\Sigma^{-1}_{\set{S}\set{S}}\Sigma_{\set{S}\set{A}'}$ for all $\set{A},\set{A}'\subset\set{X}$ and $\Lambda$ is a block-diagonal matrix constructed from the $N$ diagonal blocks of $\Sigma_{\set{D}\set{D}|\set{S}}$, each of which is a matrix $\Sigma_{\set{D}_n\set{D}_n |\set{S}}$ for $n=1, \cdots, N$ where $\set{D} = \bigcup_{n=1}^N\set{D}_n$. 
Also, unlike SoD, the support set $\set{S}$ does not have to be observed.
The covariance matrix $\Sigma_{\set{D}\set{D}}$ in \eqref{mean} and \eqref{variance} is approximated by a reduced-rank matrix $\Gamma_{\set{D}\set{D}}$ summed with the resulting sparsified residual matrix $\Lambda$ in \eqref{PITCmean} and \eqref{PITCvar}.
So, computing either $\mu^{\text{PITC}}_{x| \set{D}}$ \eqref{PITCmean} or $\sigma^{\text{PITC}}_{xx| \set{D}}$ \eqref{PITCvar}, which requires inverting 
 the approximated covariance matrix $\Gamma_{\set{D}\set{D}} + \Lambda$, incurs $\set{O}(|\set{D}|(|\set{S}|^2+(|\set{D}|/N)^2))$ time and $\set{O}(|\set{S}|^2+(|\set{D}|/N)^2)$ memory.
The sparse \emph{fully independent training conditional} (FITC) approximation method is a special case of PITC where $\Lambda$ is a diagonal matrix constructed from $\sigma_{x'x'|\set{S}}$ for all $x'\in\set{D}$ (i.e., $N=|\set{D}|$). FITC is previously employed by \citeauthor{Ko2008}~\shortcite{Ko2008} to speed up the learning of observation model with prior training data. 
But, the time incurred by PITC or FITC grows with increasing size of data. So, it is computationally impractical to use them directly to repeatedly train the observation model at each filtering step for persistent localization.
%
\subsection{Bayes Filters}
\label{sect:bf}
%
A Bayes filter is a probabilistic state estimation framework that repeatedly updates the belief of a robot's location/state by conditioning on its control actions performed and field measurements taken so far.
Formally, 
let the robot's control action performed, its location visited and observed, and the corresponding realized field measurement taken at time/filtering step $t$ be denoted by $u_t$, $x_t$, and $z_{t}$\footnote{The field measurement $z_{t}$ is indexed by time step $t$ instead of the corresponding location $x_t$ since $x_t$ is not known to the robot.}, respectively.
To estimate the robot's location, a belief $b(x_t)\triangleq p(x_t|u_{1:t}, z_{{1:t}})$ is maintained over all its possible locations $x_t$ where
$u_{1:t}\triangleq (u_1,\ldots,u_t)^\top$ and 
$z_{{1:t}}\triangleq (z_1,\ldots,z_t)^\top$
denote, respectively, column vectors of past control actions performed
and realized field measurements taken by the robot
up until time step $t$. 
To track such a belief, after the robot has performed an action $u_t$ and taken a realized measurement $z_{t}$ at each time step $t$, the Bayes filter updates the prior belief $b(x_{t-1})$ of the robot's location to the posterior belief 
$b(x_t) = \beta p(z_{t} | x_t)\int p(x_t | u_t, x_{t-1}) b(x_{t-1}) \text{d} x_{t-1}$
where $1/\beta$ is a normalizing constant,
$p(x_t | u_t, x_{t-1})$ is a \emph{motion model} representing the probability of the robot moving from locations $x_{t-1}$ to $x_t$ after performing action $u_t$,
and $p(z_{t} | x_t)$ is an \emph{observation model} describing the likelihood of taking realized measurement $z_{t}$ at location $x_t$.

To preserve efficiency, the Bayes filter imposes a Markov property on the observation model: Given the robot's current location $x_t$, its current measurement $z_{t}$ is conditionally independent of past actions $u_{1:t}$ and measurements $z_{{1:t-1}}$: \vspace{-1mm}
\begin{equation}
 p(z_{t} | x_t, u_{1:t}, z_{{1:t-1}}) = p(z_{t} | x_t)\ . 
\vspace{-1mm}
\label{markov} 
\end{equation}
In other words, the robot's past actions performed and  measurements taken  during its exploration and localization are not exploited for learning the observation model. 
This is conventionally assumed by existing works either representing the observation model using a parametric model with known parameters \cite{Thrun2005} or training it offline using prior training data. 
The disadvantages of the former are extensively discussed by \citeauthor{Ko2008}~\shortcite{Ko2008} while that of the latter are already detailed in Section~\ref{sect:intro}.

In the case of multiple fields (say, $M$ of them), let $z^m_{t}$ denote the realized measurement taken from field $m$ at location $x_t$ for $m=1,\ldots,M$.
Then, the observation model becomes 
$
p(z^1_{t},\ldots,z^M_{t} | x_t) = \prod^M_{m=1} p(z^m_{t} | x_t)
$
such that the equality follows from an assumption of independence of measurements between fields to ease computations.
\vspace{-0.4mm}
\section{Online Sparse GP Observation Model}
\label{sect:gplocalize}
%
In contrast to existing works 
discussed in Section~\ref{sect:intro}, our GP-Localize algorithm does not need to impose the restrictive Markov property \eqref{markov} on the observation model, 
which can then be derived by marginalizing out the random locations visited and observed by the robot up until time step $t-1$:
\vspace{-5mm}
\begin{equation}
\hspace{-1.8mm}
\begin{array}{l}
\displaystyle p(z_{t} | x_t, u_{1:t}, z_{{1:t-1}}) \\
\displaystyle=\hspace{-0.5mm}\eta\hspace{-1mm} \int\hspace{-1mm} b(x_0)\hspace{-0.5mm} \prod_{i=1}^{t} p(x_{i} | u_{i},x_{i-1})  p(z_{t}|x_t, x_{1:t-1}, z_{{1:t-1}}) \text{d} x_{0:t-1} \vspace{-18mm}
\end{array} 
\label{fgp} 
\vspace{11mm}
\end{equation}
where $1/\eta = p(x_t|u_{1:t},z_{{1:t-1}})$ is a normalizing constant, $b(x_0)=p(x_0)$ is the belief of the robot's initial location at time step $0$,
$x_{1:t-1}\triangleq\{x_1,\ldots,x_{t-1}\}$ denotes a set of locations visited and observed by the robot up until time step $t-1$,
and $p(z_{t}|x_t, x_{1:t-1}, z_{{1:t-1}})= \mathcal{N}(\mu_{x_t|x_{1:t-1}}, \sigma_{x_t x_t|x_{1:t-1}})$ is a Gaussian predictive distribution provided by the GP (Section~\ref{sect:gp}).
The derivation of \eqref{fgp} is in\if\myproof1 Appendix~\ref{derivation}\fi\if\myproof0 \cite{AA13}\fi.

To make computations tractable but not constrain the type of motion model that can be specified, the observation model \eqref{fgp} is approximated using Monte Carlo integration:\vspace{-2mm}
\begin{equation}
\hspace{-0mm}
 p(z_{t} | x_t, u_{1:t}, z_{{1:t-1}})\hspace{-0mm} \approx\hspace{-0mm} \frac{1}{C} \sum_{c=1}^C p(z_{t}|x_t, x^c_{1:t-1}, z_{{1:t-1}}) \hspace{-1.04mm}\vspace{-1mm}
\label{monte}
\end{equation}
where $x^c_{1:t-1}$ denotes a $c$-th sample path simulated by first drawing the robot's initial location $x^c_0$ from $b(x_0)$ and then sampling $x^c_i$ from motion model $p(x_i|u_i,x^c_{i-1})$ for $i=1,\ldots,t-1$
given its past actions $u_{1:t-1}$ 
while ensuring $p(x_t|u_t,x^c_{t-1})>0$, as observed in \eqref{fgp}.
For a practical implementation, instead of re-simulating the entire sample paths (hence, incurring linear time in $t$) at each time step,
each $c$-th sample path is incrementally updated from $x^c_{1:t-2}$ (i.e., obtained in previous time step) to $x^c_{1:t-1}$ (i.e., needed in current time step)
by including $x^c_{t-1}$ sampled from motion model $p(x_{t-1}|u_t,x^c_{t-2})$ without accounting for motion constraint $p(x_{t}|u_{t},x^c_{t-1})>0$.
As a result, the time spent in 
incrementally updating the $C$ sample paths at each time step is independent of $t$.
To mitigate the effect of ignoring the constraint, we introduce a strategy in Remark $3$ after Theorem~\ref{woohoo} 
that exploits a structural property of our proposed online sparse GP.
In practice, such an implementation yields considerable time savings (i.e., time independent of $t$) and does not result in poor localization performance empirically (Section~\ref{sect:expt}).


The scalability of our GP-Localize algorithm therefore depends on whether the Gaussian predictive probability $p(z_{t}|x_t, x^c_{1:t-1}, z_{{1:t-1}})$ in \eqref{monte} 
can be derived efficiently.
Computing it with full GP, PITC, or FITC (Section~\ref{sect:bkgd}) directly
incurs, respectively, 
$\set{O}(t^3)$, $\set{O}(t(|\set{S}|^2 + (t/N)^2))$, and $\set{O}(t|\set{S}|^2)$ time.
Since $t$ is expected to be large for persistent localization, it is computationally impractical to use these offline full GP and sparse GP approximation methods to repeatedly train the observation model at each filtering step.
Even when the online GP proposed by \citeauthor{Csato02}~\shortcite{Csato02} is used, it still incurs 
quadratic time in $t$ per filtering step.
In the following subsection, we will propose an online sparse GP
that can achieve constant time (i.e., independent of $t$) at each filtering step $t$.
\vspace{-1mm}
%
%
%
%
\subsection{Online Sparse GP Approximation}
\label{sect:osgp}
The key idea underlying our proposed online sparse GP is to summarize the newly gathered data/observations at regular time intervals/slices, assimilate the summary information of the new data with that of all the previously gathered data/observations,
and then exploit the resulting assimilated summary information to compute the Gaussian predictive probability $p(z_{t}|x_t, x^c_{1:t-1}, z_{{1:t-1}})$ in \eqref{monte}.
The details of our proposed online sparse GP will be described next.

Let each time slice $n$ span time/filtering steps $(n-1)\tau+1$ to $n\tau$ for some user-defined slice size $\tau\in\mathbb{Z}^{+}$ and the number of time slices available thus far up until time step $t$ be denoted by $N$ (i.e., $N\tau<t$).
\begin{definition}[Slice Summary] 
Given a support set $\set{S}\subset\set{X}$ common to all $C$ sample paths, 
the subset $\set{D}_n \triangleq x_{(n-1)\tau+1:n\tau}^{c}$ of the $c$-th sample path $x_{1:t-1}^{c}$ simulated during the time slice $n$,
and the column vector $z_{\set{D}_n}=z_{(n-1)\tau+1:n\tau}$ of corresponding realized measurements taken by the robot, the slice summary of time slice $n$ is defined as a tuple
$({\mu}^n_{\circled{s}}, {\Sigma}^n_{\circled{s}})$ for $n=1,\ldots, N$ where
\vspace{-1mm}
\begin{equation*}
{\mu}^n_{\circled{s}} \triangleq \Sigma_{\set{S}\set{D}_n}\Sigma^{-1}_{\set{D}_n \set{D}_n|\set{S}}(z_{\set{D}_n}-\mu_{\set{D}_n})
\vspace{-1mm}
%
\end{equation*}
%
\begin{equation*}
{\Sigma}^n_{\circled{s}} \triangleq
\Sigma_{\set{S}\set{D}_n}\Sigma_{\set{D}_n \set{D}_n|\set{S}}^{-1}\Sigma_{\set{D}_n \set{S}}
%
\end{equation*}
such that $\mu_{\set{D}_n}$ is defined in a similar manner as $\mu_\set{D}$ in \eqref{mean} and $\Sigma_{\set{D}_n \set{D}_n|\set{S}}$ is a posterior covariance matrix with components $\sigma_{xx'|\set{S}}$ for all $x,x' \in \set{D}_n$, each of which is defined in a similar way as \eqref{pcovsod}. 
\label{def:ls}
\end{definition}
%
\emph{Remark}. The support set $\set{S}\subset\set{X}$ of locations does not have to be observed because the slice summary is independent of 
$z_\set{S}$. 
So, the support set $\set{S}$ can be selected \emph{prior} to exploration and localization from 
$\set{X}$ 
using an offline greedy active learning algorithm such as \cite{Krause08}.
%
\begin{definition}[Assimilated Summary]
Given 
$({\mu}^n_{\circled{s}}, {\Sigma}^n_{\circled{s}})$,
the assimilated summary $({\mu}^n_{\circled{a}}, {\Sigma}^n_{\circled{a}})$ of time slices $1$ to $n$ is updated from the assimilated summary $({\mu}^{n-1}_{\circled{a}}, {\Sigma}^{n-1}_{\circled{a}})$ of time slices $1$ to $n-1$ using
$
{\mu}^n_{\circled{a}} \triangleq {\mu}^{n-1}_{\circled{a}} + {\mu}^n_{\circled{s}} 
$
and
$
{\Sigma}^n_{\circled{a}}\triangleq {\Sigma}^{n-1}_{\circled{a}} + {\Sigma}^n_{\circled{s}} 
$ 
for $n=1,\ldots, N$ where
$
{\mu}^0_{\circled{a}} \triangleq 0
$
and
$
{\Sigma}^0_{\circled{a}} \triangleq \Sigma_{\set{S}\set{S}}
$.
\label{as}
\end{definition}
\emph{Remark} $1$. After constructing and assimilating
$({\mu}^n_{\circled{s}}, {\Sigma}^n_{\circled{s}})$ 
with $({\mu}^{n-1}_{\circled{a}}, {\Sigma}^{n-1}_{\circled{a}})$ to form $({\mu}^n_{\circled{a}}, {\Sigma}^n_{\circled{a}})$,
$\set{D}_n = x_{(n-1)\tau+1:n\tau}^{c}$, 
$z_{\set{D}_n}= z_{(n-1)\tau+1:n\tau}$, and 
$({\mu}^n_{\circled{s}}, {\Sigma}^n_{\circled{s}})$
(Definition~\ref{def:ls}) are no longer needed and can be removed from memory.
As a result, at time step $t$ where $N\tau+1\leq t \leq (N+1)\tau$, only 
$({\mu}^N_{\circled{a}}, {\Sigma}^N_{\circled{a}})$, 
$x_{N\tau+1:t-1}^c$, and 
$z_{N\tau+1:t-1}$ have to be kept in memory, thus requiring only constant memory (i.e., independent of $t$).\vspace{1mm}

\noindent
\emph{Remark} $2$. The slice summaries are constructed and assimilated at a regular time interval of $\tau$, specifically, at time steps $N\tau+1$ for $N\in\mathbb{Z}^{+}$.
%
\begin{theorem}
Given 
$\set{S}\subset\set{X}$ 
and $({\mu}^N_{\circled{a}}, {\Sigma}^N_{\circled{a}})$, 
our online sparse GP computes a Gaussian predictive distribution $p(z_{t}|x_t, {\mu}^N_{\circled{a}}, {\Sigma}^N_{\circled{a}})= \set{N}(\widetilde{\mu}_{x_t}, \widetilde{\sigma}_{x_t x_t})$
of the measurement at any 
location $x_t\in\set{X}$ 
at time step $t$ (i.e., $N\tau+1\leq t \leq (N+1)\tau$)
with the following posterior mean and variance: \vspace{-1mm}
\begin{equation}
 \widetilde{\mu}_{x_t} \triangleq \mu_{x_t} + \Sigma_{x_t \set{S}}\left({\Sigma}^N_{\circled{a}}\right)^{-1}{\mu}^N_{\circled{a}} 
 \label{pmeanold}
  \end{equation}
  \vspace{-3.6mm}
  \begin{equation}
 \widetilde{\sigma}_{x_t x_t} \triangleq \sigma_{x_t x_t} - \Sigma_{x_t \set{S}}\left(\Sigma^{-1}_{\set{S}\set{S}} - \left({\Sigma}^N_{\circled{a}}\right)^{-1}\right)\Sigma_{\set{S} x_t}\ . 
 \label{pvarold}
 \end{equation}
If $t=N\tau+1$, $\widetilde{\mu}_{x_t}=\mu^{\text{\em PITC}}_{x_t| x^c_{1:t-1}}$ and $\widetilde{\sigma}_{x_t x_t}=\sigma^{\text{\em PITC}}_{x_t x_t| x^c_{1:t-1}}$. 
\label{woohoo}
\end{theorem}
Its proof is given in\if\myproof1 Appendix~\ref{derivation2}\fi\if\myproof0 \cite{AA13}\fi. \vspace{1mm}

\noindent
\emph{Remark} $1$. Theorem~\ref{woohoo} implies that our proposed online sparse GP is in fact equivalent to an online learning formulation/variant of the offline PITC (Section~\ref{sect:sgp}).
Supposing $\tau<|\set{S}|$, the $\set{O}(t|\set{S}|^2)$ time incurred by offline PITC to compute 
$p(z_{t}|x_t, x^c_{1:t-1}, z_{{1:t-1}})$ in \eqref{monte}
can then be reduced to $\set{O}(\tau|\set{S}|^2)$ time (i.e., time independent of $t$) incurred by our online sparse GP at time steps $t=N\tau+1$ for $N\in\mathbb{Z}^{+}$ when slice summaries are constructed and assimilated. Otherwise, our online sparse GP only incurs $\set{O}(|\set{S}|^2)$ time per time step.\vspace{1mm}
%
%

\noindent
\emph{Remark} $2$. The above equivalence result allows the structural property of our online sparse GP to be elucidated using that of offline PITC:
The measurements
$Z_{\set{D}_1},\ldots,Z_{\set{D}_N}, Z_{x_t}$ 
between different time slices are assumed to be conditionally independent given $Z_{\set{S}}$. 
Such an assumption enables the data gathered during each time slice to be summarized independently of that in other time slices.
Increasing slice size $\tau$ (i.e., less frequent assimilations of larger slice summaries) relaxes this conditional independence assumption (hence, potentially improving the fidelity of the resulting observation model), but 
incurs more time at time steps when slice summaries are constructed and assimilated (see Remark $1$).
\vspace{1mm}
%

\noindent
\emph{Remark} $3$. Recall (see paragraph after \eqref{monte}) that the motion constraint $p(x_{t}|u_{t},x^c_{t-1})>0$ is not accounted for when sampling $x^c_{t-1}$ from motion model $p(x_{t-1}|u_t,x^c_{t-2})$ at each time step $t$.
To mitigate the effect of ignoring the constraint,
at time steps $t=N\tau+2$ for $N\in\mathbb{Z}^{+}$,
we draw ${x}'_{N\tau}$ from the particle-based belief $b(x_{N\tau})$ maintained in our experiments (Section~\ref{sect:expt})
and use it (instead of $x^{c}_{N\tau}$) for sampling $x^c_{N\tau+1}$ from motion model $p(x_{N\tau+1}|u_i,{x}'_{N\tau})$.
Doing this at a regular time interval of $\tau$ reduces the deviation of the simulated sample paths from the particle-based beliefs updated at each time step and consequently allows the sample paths to satisfy the motion constraint more often, especially when $\tau$ is small. Such a strategy may cause the sampled $x^c_{N\tau+1}$ not to be located close to $x^{c}_{N\tau}$ (hence, their corresponding realized measurements are less spatially correlated) since $x^c_{N\tau+1}$ is not sampled from motion model $p(x_{N\tau+1}|u_i,x^{c}_{N\tau})$. But, this occurs at a lower frequency of $1/\tau$, as compared to not considering the motion constraint at every time step.
Furthermore, this fits well with the structural property of our online sparse GP that assumes $Z_{N\tau}$ and $Z_{N\tau+1}$ (or, more generally, $Z_{\set{D}_N}$ and $Z_{x_t}$) to be conditionally independent.\vspace{1mm}

\noindent
\emph{Remark} $4$. Since offline PITC generalizes offline FITC (Section~\ref{sect:sgp}), our online sparse GP generalizes
the online learning variant of FITC (i.e., $\tau=1$) \cite{Csato02}\footnote{\citeauthor{SnelsonThesis07}~\shortcite{SnelsonThesis07} 
pointed out that the sparse online GP of \citeauthor{Csato02}~\shortcite{Csato02} is an online learning variant of offline FITC.}.\vspace{1mm}

When $N\tau+1<t\leq(N+1)\tau$ (i.e., before the next slice summary of time slice $N+1$ is constructed and assimilated),
the most recent observations (i.e., $\set{D}'\triangleq x_{N\tau+1:t-1}^c$ and $z_{\set{D}'} = z_{N\tau+1:t-1}$), which are often highly informative, are not used to update $\widetilde{\mu}_{x_t}$ \eqref{pmeanold} and $\widetilde{\sigma}_{x_t x_t}$ \eqref{pvarold}.
This will hurt the localization performance, especially when $\tau$ is large and the robot is localizing in an unexplored area with little/no observations; the field within this area thus cannot be predicted well with the current assimilated summary. To resolve this, we exploit incremental update formulas of Gaussian posterior mean and variance\if\myproof1 (Appendix~\ref{sect:derivation3}) \fi\if\myproof0 \cite{AA13} \fi to update $\widetilde{\mu}_{x_t}$ and $\widetilde{\sigma}_{x_t x_t}$ with the most recent observations,  thereby yielding a Gaussian predictive distribution
$p(z_{t}|x_t, {\mu}^N_{\circled{a}}, {\Sigma}^N_{\circled{a}},\set{D}', z_{\set{D}'})= \set{N}(\widetilde{\mu}_{x_t|\set{D}'}, \widetilde{\sigma}_{x_t x_t|\set{D}'})$
where \vspace{-1mm}
\begin{equation}
\widetilde{\mu}_{x_t|\set{D}'} \triangleq \widetilde{\mu}_{x_t} + \widetilde{\Sigma}_{x_t \set{D}'}\widetilde{\Sigma}^{-1}_{\set{D}'\set{D}'}\left(z_{\set{D}'} - \widetilde{\mu}_{\set{D}'}\right) \vspace{-0.6mm}
 \label{pmeannew}
 \end{equation}
\begin{equation}
\widetilde{\sigma}_{x_t x_t|\set{D}'} \triangleq \widetilde{\sigma}_{x_t x_t} - \widetilde{\Sigma}_{x_t \set{D}'}\widetilde{\Sigma}^{-1}_{\set{D}'\set{D}'}\widetilde{\Sigma}_{\set{D}' x_t} \vspace{-0mm}
\label{pvarnew}
 \end{equation}
such that $\widetilde{\mu}_{\set{D}'}$ is a column vector with mean components $\widetilde{\mu}_{x}$ (i.e., defined similarly to \eqref{pmeanold}) 
for all $x\in \set{D}'$,
$\widetilde{\Sigma}_{x_t\set{D}'}$ is a row vector with covariance components $\widetilde{\sigma}_{x_t x}$ 
(i.e., defined similarly to \eqref{pvarold}) 
for all $x\in \set{D}'$, $\widetilde{\Sigma}_{\set{D}'x_t}$ is the transpose of $\widetilde{\Sigma}_{x_t\set{D}'}$, and $\widetilde{\Sigma}_{\set{D}'\set{D}'}$ is a matrix with covariance components $\widetilde{\sigma}_{xx'}$ 
(i.e., defined similarly to \eqref{pvarold}) 
for all $x,x' \in \set{D}'$.
%
\begin{theorem}
Computing $p(z_{t}|x_t, {\mu}^N_{\circled{a}}, {\Sigma}^N_{\circled{a}},\set{D}', z_{\set{D}'})$ (i.e., \eqref{pmeannew} and \eqref{pvarnew}) incurs $\set{O}(\tau|\set{S}|^2)$ time at time steps $t=N\tau+1$ for $N\in\mathbb{Z}^{+}$ and $\set{O}(|\set{S}|^2)$ time otherwise.
It requires $\set{O}(|\set{S}|^2)$ memory at each time step.
\label{timespace}
\end{theorem}
Its proof is given in\if\myproof1 Appendix~\ref{sect:complexity}\fi\if\myproof0  \cite{AA13}\fi. Theorem~\ref{timespace} indicates that our online sparse GP incurs constant time and memory (i.e., independent of $t$) per time step.
 %
%
%
%
\section{Experiments and Discussion}
\label{sect:expt}
This section evaluates the localization performance, time efficiency, and scalability of our GP-Localize algorithm empirically through simulated experiments with three real-world datasets: 
(a) \emph{Wireless signal strength} (WSS) (signal-to-noise ratio) data \cite{Guestrin07} produced by $6$ WiFi access points (APs) and measured at over $200$ locations throughout the fifth floor of Wean Hall in Carnegie Mellon University (Fig.~\ref{fig:wifi-traj}, Section~\ref{wssf}),
(b) \emph{indoor environmental quality} (IEQ) (i.e., temperature ($\hspace{-0.5mm}\,^{\circ}\mathrm{F}$) and light (Lux)) data \cite{Guestrin04} measured by $54$ sensors deployed in the Intel Berkeley Research lab (Fig.~\ref{fig:ieq-traj}, Section~\ref{ieq}),
(c) \emph{urban traffic speeds} (UTS) (km/h) data \cite{LowUAI12,LowUAI13} measured at $775$ road segments (including highways, arterials, slip roads, etc.) of an urban road network in Tampines area, Singapore during evening peak hours on April $20$, $2011$ with a mean speed of $47.6$ km/h and a standard deviation of $20.5$ km/h (Fig.~\ref{fig:traffic}a, Section~\ref{traffic}),
and (d) a real Pioneer $3$-DX mobile robot (i.e., mounted with a weather board) experiment on a trajectory of about $280$~m in the \emph{Singapore-MIT Alliance for Research and Technology Future Urban Mobility} (SMART FM) IRG office/lab
gathering 
$561$ relative light ($\%$) 
observations/data for GP localization (Fig.~\ref{p3dxtraj}, Section~\ref{p3dx}).
Different from the $2$-dimensional spatial domains of the WSS, IEQ, and light fields,
each road segment of the urban road network is specified by a $5$-dimensional vector of features: length, number of lanes, speed limit, direction, and time. 
The UTS field is modeled using a relational GP (previously developed in \cite{LowUAI12}) whose correlation structure can exploit both the road segment features and road network topology information.
The hyperparameters 
of each GP modeling a different field are learned using the data via maximum likelihood estimation \cite{Rasmussen2006}.
Our GP-Localize algorithm is implemented using an odometry motion model\footnote{\label{const}Due to lack of space, an interested reader is referred to \cite{Thrun2005} for the technical details of the odometry motion model and particle filter.}, our online sparse GP (i.e., setting $\tau=10$ and $|\set{S}|=40$) for representing the observation model (Section~\ref{sect:gplocalize}), and a particle filter\cref{const} of $400$ particles for representing the belief of the robot's location. 
The number $C$ of sample paths in \eqref{monte} is set to $400$ for all experiments.
For the simulated experiments with the WSS and IEQ data, the control actions (i.e., odometry information) are generated using the realistic Pioneer mobile robot module in Player/Stage simulator 
\cite{Gerkey_2003} 
and the measurements taken along the generated trajectory of $421$ ($336$) time steps from the WSS (IEQ) fields shown in Fig.~\ref{fig:wifi-traj} (Fig.~$2$) are the Gaussian predictive/posterior means \eqref{mean} of each full GP modeling a separate field trained using the data.
For the simulated experiment with the UTS data, the control actions of the mobile probe vehicle are assumed not to be known; its transition probability of moving from one road segment to another can be learned from vehicle route data using the hierarchical Bayesian nonparametric approach of \citeauthor{LowIAT12}~\shortcite{LowIAT12}. The measurements taken along its generated trajectory of $370$ time steps from the UTS field are shown in Fig.~\ref{fig:traffic}.

%

The localization performance/error (i.e., distance between the robot's estimated and true locations) and scalability of our GP-Localize algorithm is compared to that of two sparse GP localization algorithms: (a) The \emph{SoD-Truncate} method uses $|\set{S}|=10$ most recent observations (i.e., compared to $|\set{D}'|<\tau=10$ most recent observations considered by our online sparse GP besides the assimilated summary) as training data at each filtering step while (b) the \emph{SoD-Even} method uses $|\set{S}|=40$ observations (i.e., compared to the support set of $|\set{S}|=40$ possibly unobserved locations selected \emph{prior} to localization and exploited by our online sparse GP) evenly distributed over the time of localization.
The scalability of GP-Localize is further compared to that of GP localization algorithms employing full GP and offline PITC (Section~\ref{sect:bkgd}).
%
\subsection{Wireless Signal Strength (WSS) Fields}
\label{wssf}
\vspace{-0.4mm}
%
%
\begin{figure}
\begin{tabular}{c}
\hspace{-2mm}\includegraphics[scale=0.351]{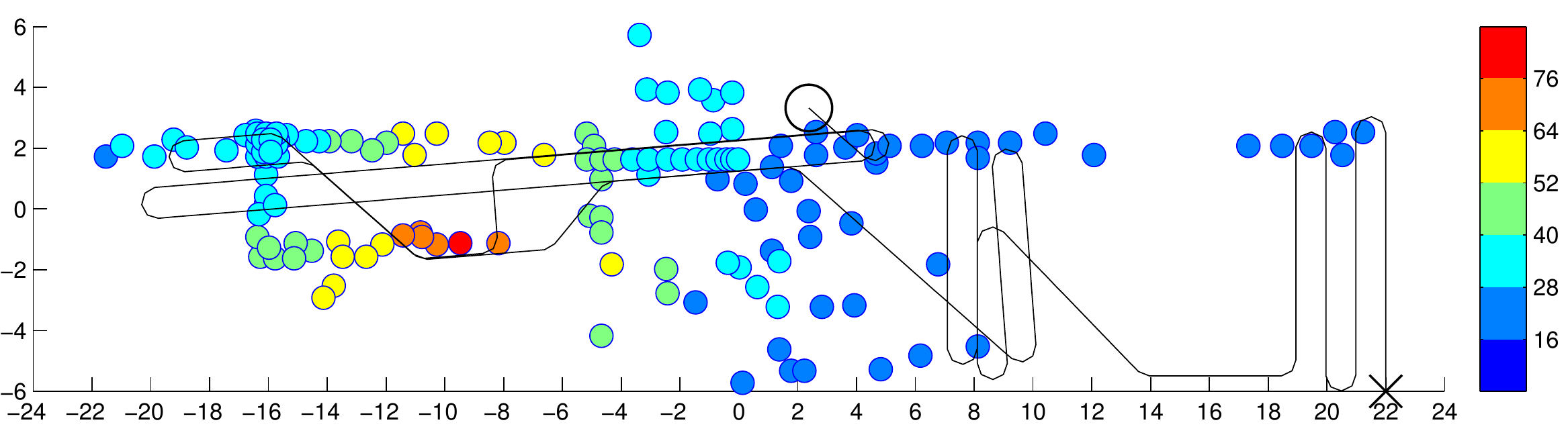} \vspace{-2mm}\\
\hspace{-2mm}{\scriptsize (a) Access point $3$}\\
\hspace{-2mm}\includegraphics[scale=0.351]{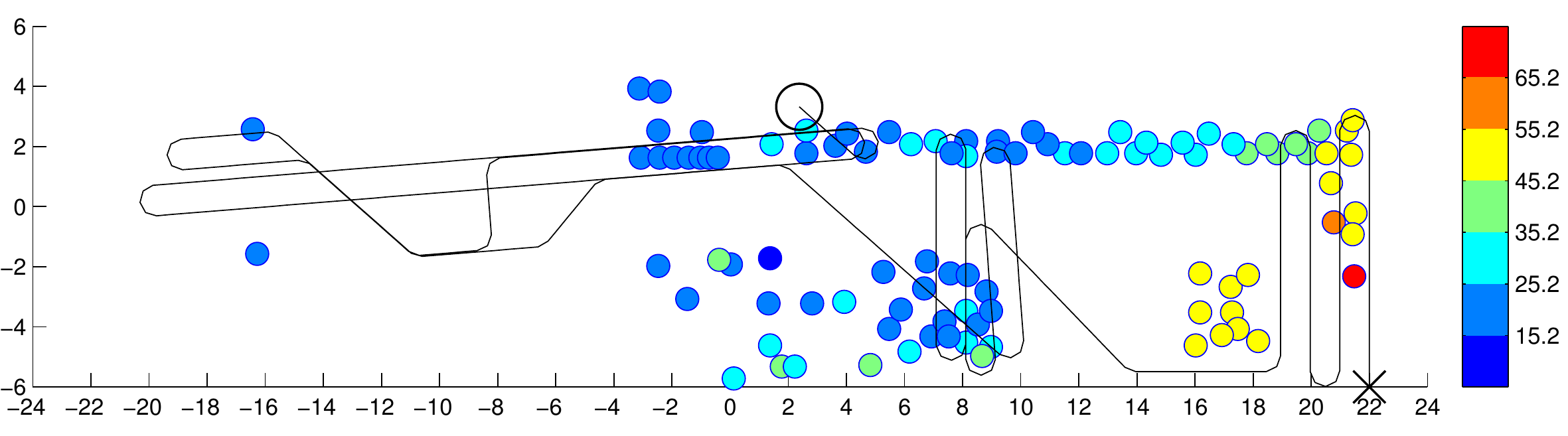} \vspace{-2mm}\\
\hspace{-2mm}{\scriptsize (b) Access point $4$}\vspace{-3mm}
\end{tabular}
\caption{WSS (signal-to-noise ratio) data produced by WiFi APs $3$ and $4$ and measured at locations denoted by small colored circles.
Robot trajectory starts at `$\times$' and ends at `$\bigcirc$'. WSS data produced by WiFi APs $1$, $2$, $5$, $6$ are shown in\if\myproof1 Appendix~\ref{fig:wss1256}\fi\if\myproof0 \cite{AA13}\fi.}\vspace{-4mm}
\label{fig:wifi-traj}
\end{figure}
%
%
\begin{table}[b]
\vspace{-4mm}
{\small
\begin{tabular}{lccccccc}
\hline
Field &\hspace{-3mm} 1 &\hspace{-2mm} 2 &\hspace{-2mm} 3 &\hspace{-2mm} 4 &\hspace{-2mm} 5\hspace{-2mm} &\hspace{-2mm} 6 &\hspace{-4mm} Multiple\\
\hline
{\bf GP-Localize} &\hspace{-3mm} 17.0 &\hspace{-2mm} 17.1 &\hspace{-2mm} 19.4 &\hspace{-2mm} 8.2 &\hspace{-2mm} 17.6 &\hspace{-2mm} 17.0 &\hspace{-4mm} 6.7 \\
SoD-Truncate &\hspace{-3mm} 25.0 &\hspace{-2mm} 23.1 &\hspace{-2mm} 23.4 &\hspace{-2mm} 21.4 &\hspace{-2mm} 21.0 &\hspace{-2mm} 22.4 &\hspace{-4mm} 20.1 \\
SoD-Even  &\hspace{-3mm} 21.2 &\hspace{-2mm} 20.5 &\hspace{-2mm} 22.9 &\hspace{-2mm} 21.9 &\hspace{-2mm} 21.5 &\hspace{-2mm} 21.5 &\hspace{-4mm} 21.8 \\
\hline
\end{tabular}
}
\vspace{-5mm}
\caption{Localization errors in single WSS fields $1$, $2$, $3$, $4$, $5$, $6$ corresponding to APs $1$, $2$, $3$, $4$, $5$, $6$ and in multiple fields.}
\label{tab:wss}
\end{table}
Table~\ref{tab:wss} shows the localization errors (no units given in WSS data) of GP-localize, SoD-Truncate, and SoD-Even averaged over all $421$ time steps and $5$ simulation runs.
%
It can be observed that GP-Localize outperforms the other two methods in every single field and in multiple fields (i.e., all $6$):
The observation model (i.e., represented by our online sparse GP) of GP-Localize can exploit the assimilated summary and the most recent observations to predict, respectively,  the fields in explored and unexplored areas better.
In contrast, SoD-Truncate performs poorly in explored areas since its predictive capability is limited by using only the most recent observations.
%
The limited observations of SoD-Even can only cover the entire area sparsely, thus producing an inaccurate observation model.

It can also be observed from Table~\ref{tab:wss} that GP-Localize achieves its largest (smallest) localization error in field $3$ ($4$):
Fig.~\ref{fig:wifi-traj}a shows that the robot does not explore the area on the left with highly varying measurements well enough, thus yielding an assimilated summary that is less informative to localization in this area. Though it explores the area on the right densely, the field in this area is relatively constant, hence making localization difficult.
As a result, localization error is high in field $3$.
On the other hand, Fig.~\ref{fig:wifi-traj}b shows that the robot explores the area on the right with highly varying measurements densely, thus achieving low error in field $4$.
\begin{figure}
\begin{tabular}{cc}
\hspace{-1.5mm}\includegraphics[scale=0.232]{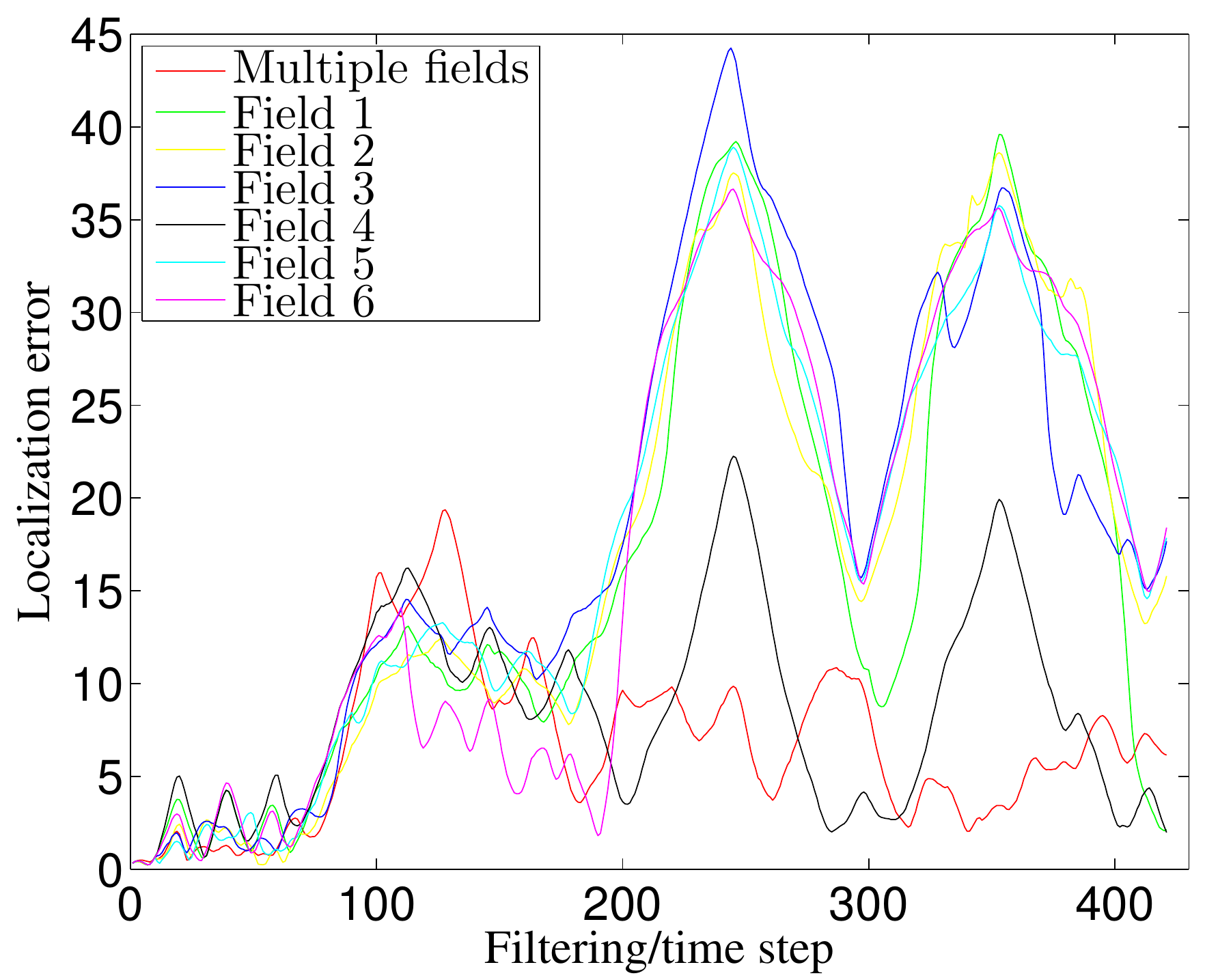} & \hspace{-5mm}
\includegraphics[scale=0.232]{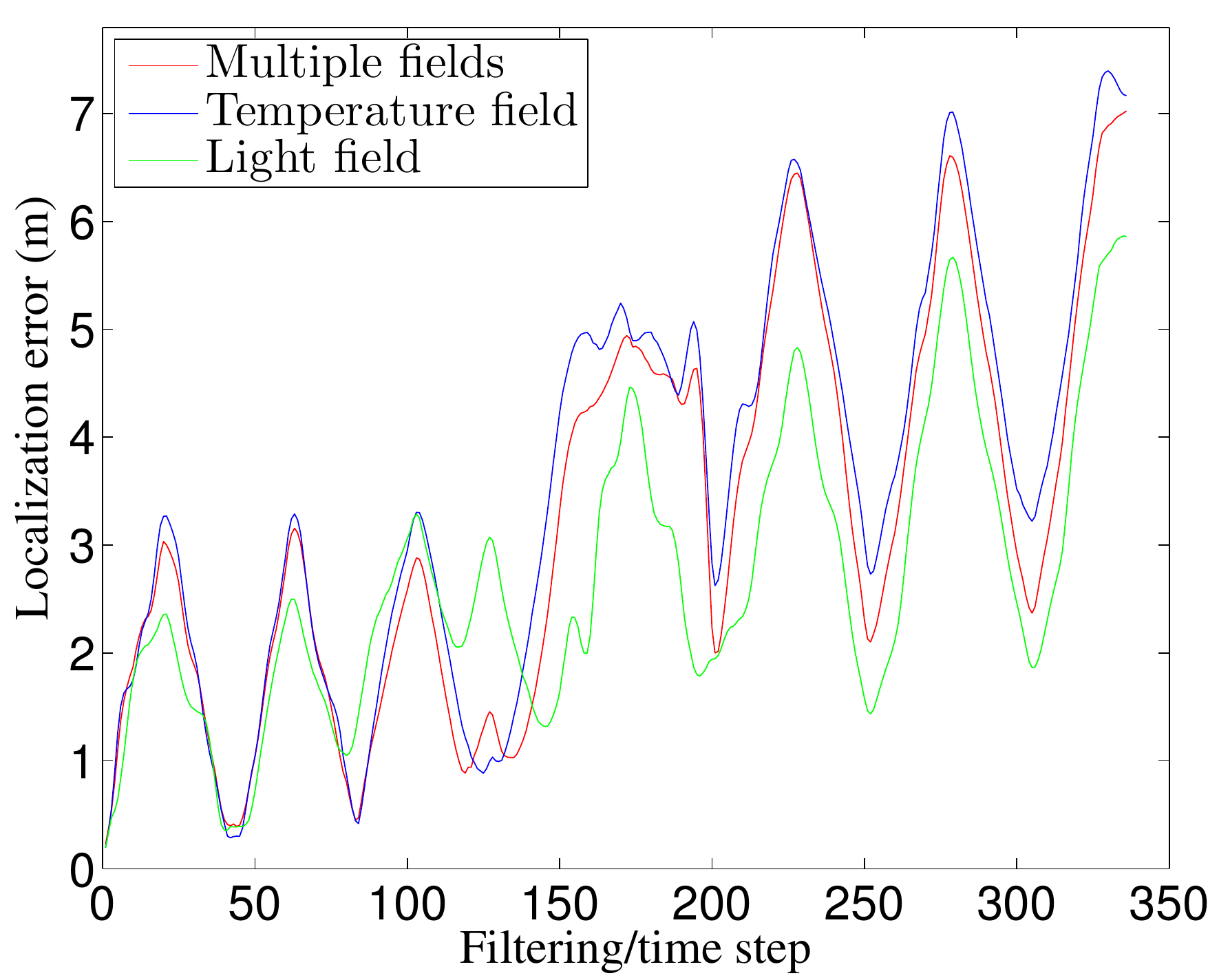} \vspace{-2mm}\\
\hspace{-1.5mm}{\scriptsize (a) WSS fields} & \hspace{-5mm}{\scriptsize (b) IEQ fields}\vspace{-4mm}
\end{tabular}
\caption{Graphs of localization error vs. no. of time steps.}\vspace{-2mm}
\label{fig:wifiieqstep}
\end{figure}


Fig.~\ref{fig:wifiieqstep}a shows the localization error of GP-Localize at each time step in every single field and in multiple fields (i.e., all $6$) averaged over $5$ runs.
It can be observed that although the error in multiple fields is not always smallest 
at each time step, it often tends to be close to (if not, lower than) the lowest error among all single fields and, more importantly, is not so high like those in single fields $1$, $2$, $3$, $5$, $6$ after $200$ time steps.
In practice, since it is usually not known which single field yields a low or high error at each time step, a more robust GP-Localize algorithm (i.e., achieved by exploiting multiple fields) is preferred.
%
\subsection{Indoor Environmental Quality (IEQ) Fields}
\label{ieq}
%
\begin{figure}
\begin{tabular}{cc}
\hspace{-2mm}\includegraphics[scale=0.252]{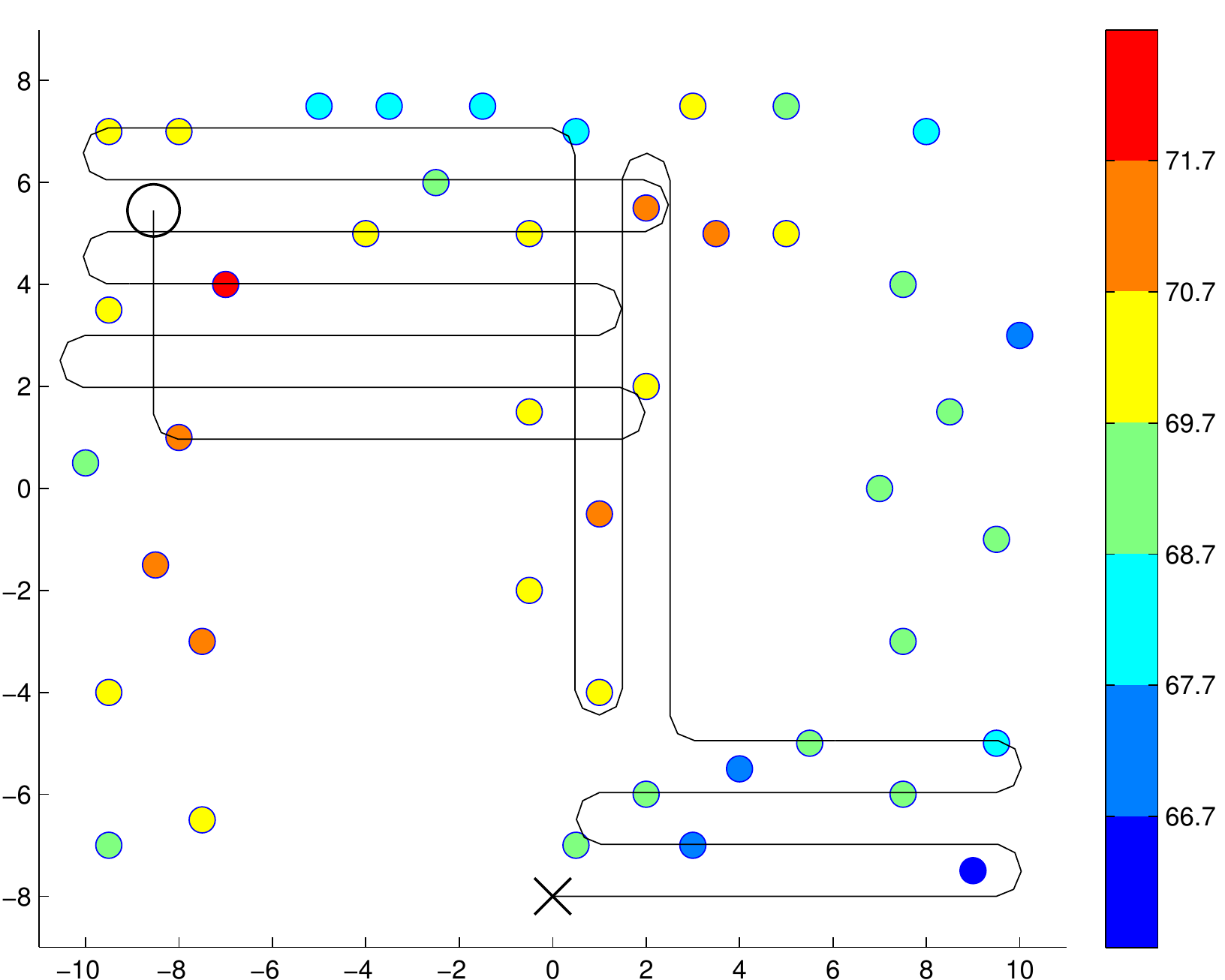} & \hspace{-4mm}\includegraphics[scale=0.252]{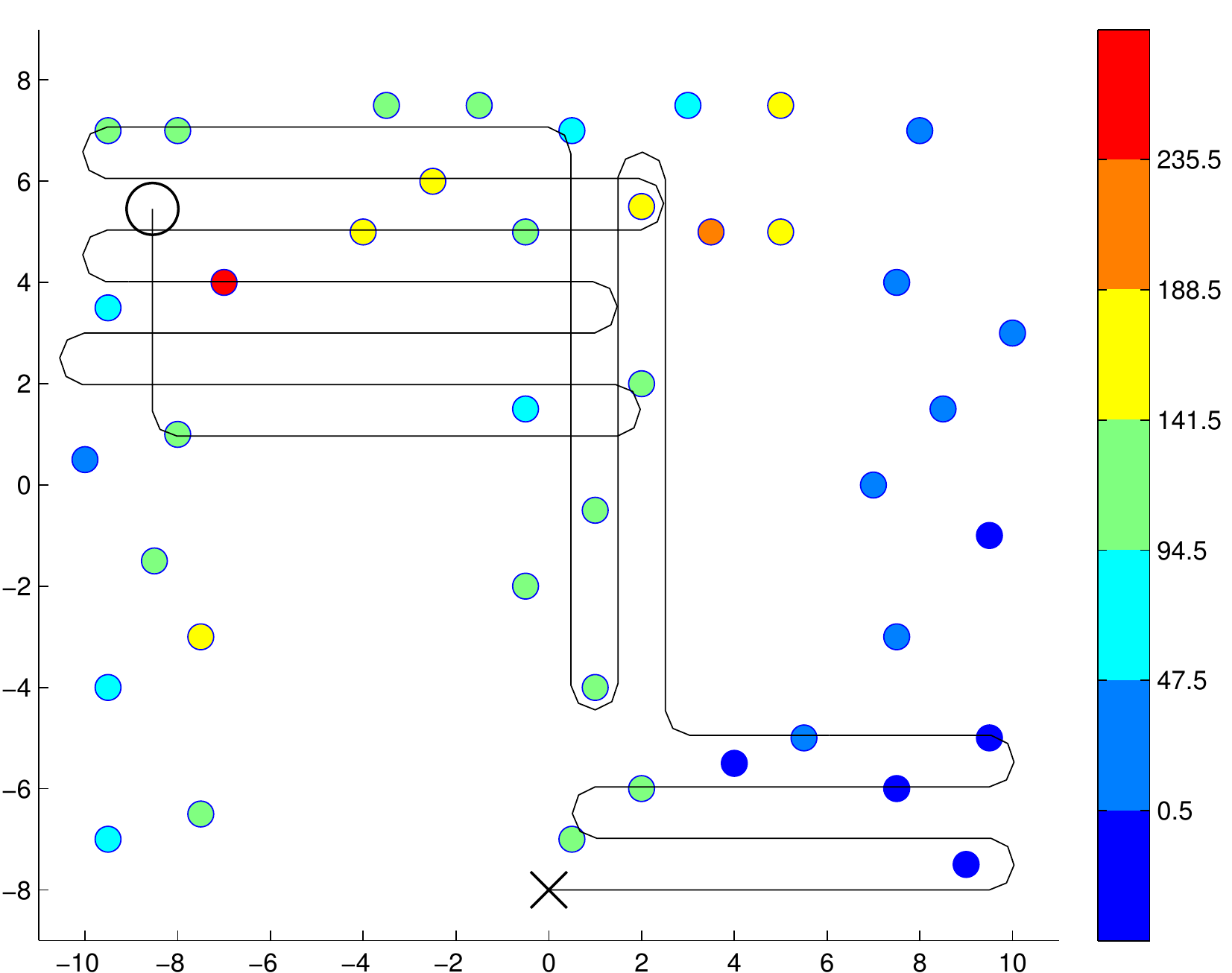} \vspace{-1mm}\\
\hspace{-2mm}{\scriptsize (a) Temperature ($\hspace{-0.5mm}\,^{\circ}\mathrm{F}$)} & \hspace{-4mm}{\scriptsize (b) Light (Lux)}\vspace{-3mm}
\end{tabular}
\caption{Temperature ($\hspace{-0.5mm}\,^{\circ}\mathrm{F}$) and light (Lux) data measured at sensor locations denoted by small colored circles. Robot trajectory starts at `$\times$' and ends at `$\bigcirc$'. Each unit along the vertical and horizontal axes is $1$~m.}\vspace{-4mm}
\label{fig:ieq-traj}
\end{figure}
\begin{table}[b]
\vspace{-4mm}
\centering
\begin{tabular}{lccc}
\hline
Field & Temperature & Light & Multiple\\
\hline
{\bf GP-Localize} & 3.5 & 2.7 & 3.2 \\
SoD-Truncate & 4.0 & 6.6 & 3.8 \\
SoD-Even  & 5.3 & 3.3 & 4.4 \\
\hline
\end{tabular}
\vspace{-3mm}
\caption{Localization errors (m) in single and multiple IEQ fields.}
\label{tab:ieq}
\end{table}
Table~\ref{tab:ieq} shows the localization errors (m) of the tested methods averaged over all $336$ time steps and $5$ simulation runs.
Similar to the observations for WSS fields (Section~\ref{wssf}), GP-Localize outperforms the other two methods in every single field and in multiple fields, as explained in the previous section.
It can also be observed that GP-Localize achieves a smaller error in the light field than in the temperature field because the measurements of the light field vary slightly more than that of the temperature field, as shown in Fig.~\ref{fig:ieq-traj}. 
Fig.~\ref{fig:wifiieqstep}b shows the error of GP-Localize at each time step in every single field and in multiple fields averaged over $5$ runs.
It can again be observed that although the error in multiple fields is not always smallest 
at each time step, 
it is often close to (if not, lower than) the lowest error among all single fields and not as high as that in the single temperature field.
Our GP-Localize algorithm exploiting multiple fields is therefore more robust in this experiment.
%
%
\subsection{Urban Traffic Speeds (UTS) Field}
\label{traffic}
\begin{figure}
\begin{tabular}{cc}
\hspace{-4mm}\includegraphics[scale=0.27]{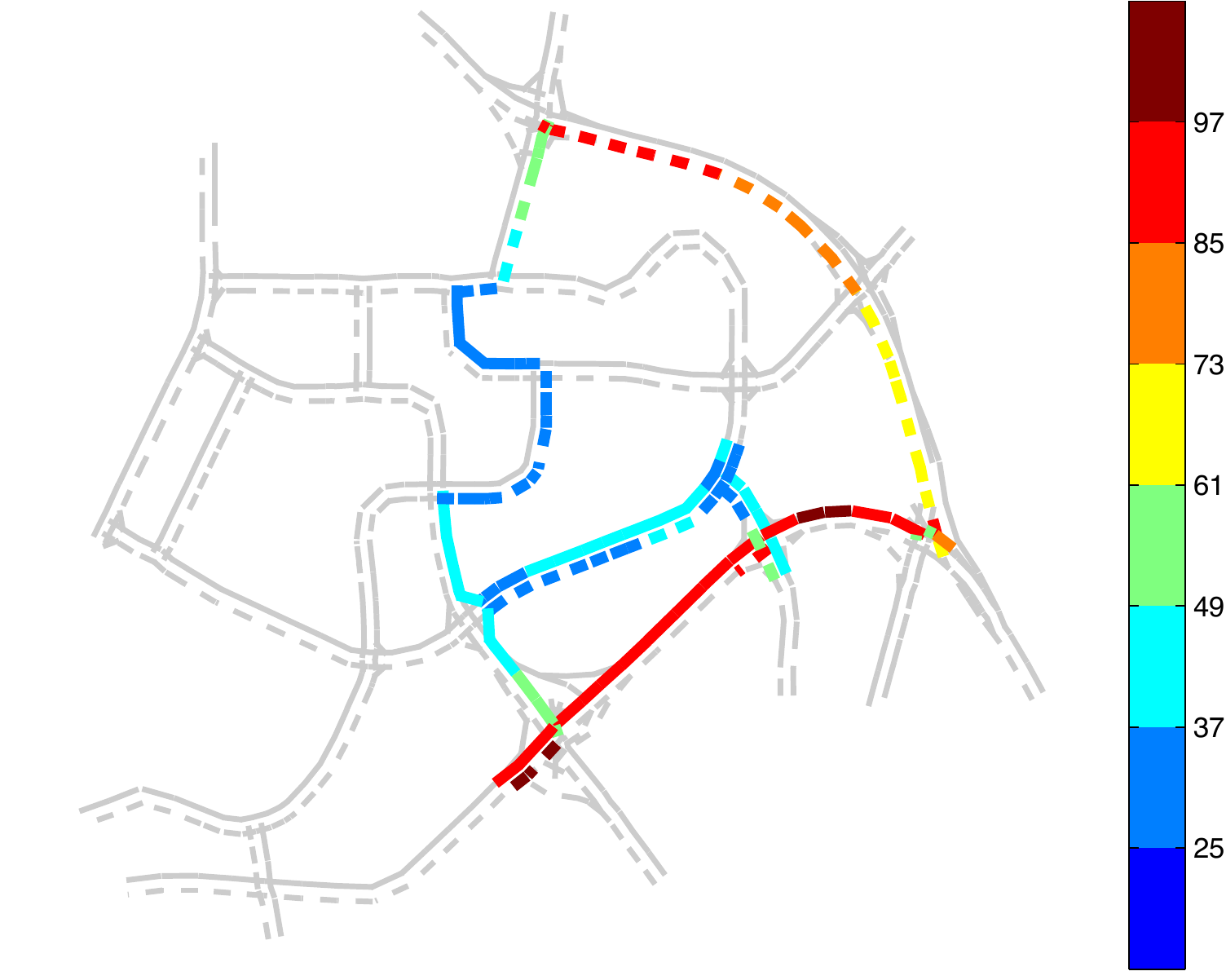} & \hspace{-3mm}\includegraphics[scale=0.23]{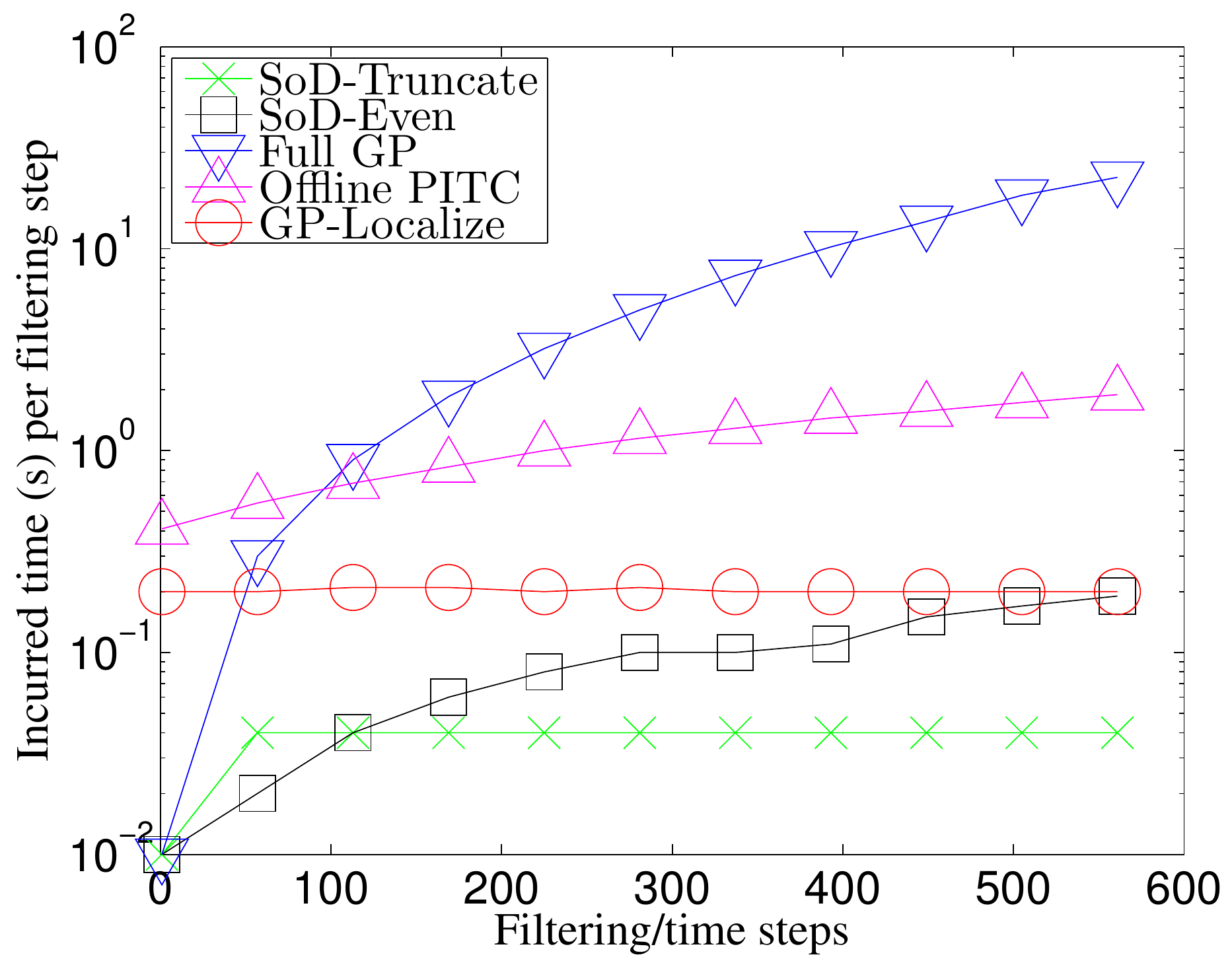}\vspace{-1mm}\\
\hspace{-4mm}{(a)} & \hspace{-3mm}{(b)} 
\vspace{-3mm}
\end{tabular}
\caption{(a) UTS (km/h) data taken along the vehicle trajectory denoted by colored road segments of the urban road network in Tampines area, Singapore, and (b) graphs of incurred time (s) per filtering step vs. no. of filtering/time steps comparing different GP localization algorithms in the real Pioneer $3$-DX mobile robot experiment.}\vspace{-1mm}
\label{fig:traffic}
\label{fig:time}
\end{figure}
For the UTS field, the localization error is defined as the geodesic (i.e., shortest path) distance between the vehicle's estimated and true residing road segments with respect to the road network topology (Fig.~\ref{fig:traffic}).
GP-Localize, SoD-Truncate, and SoD-Even achieve, respectively, localization errors of $2.8$, $7.3$, and $6.2$ road segments averaged over all $370$ time steps and $3$ simulation runs.
\subsection{Real Pioneer 3-DX Mobile Robot Experiment}
\label{p3dx}
%
\begin{figure}
\begin{tabular}{cc}
\hspace{-2mm}\includegraphics[height=35mm,width=39mm]{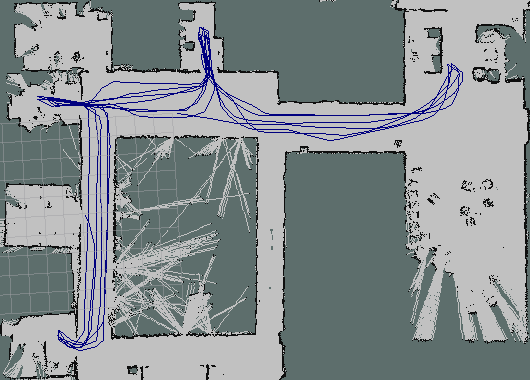}
&\hspace{-3mm}\includegraphics[height=35mm]{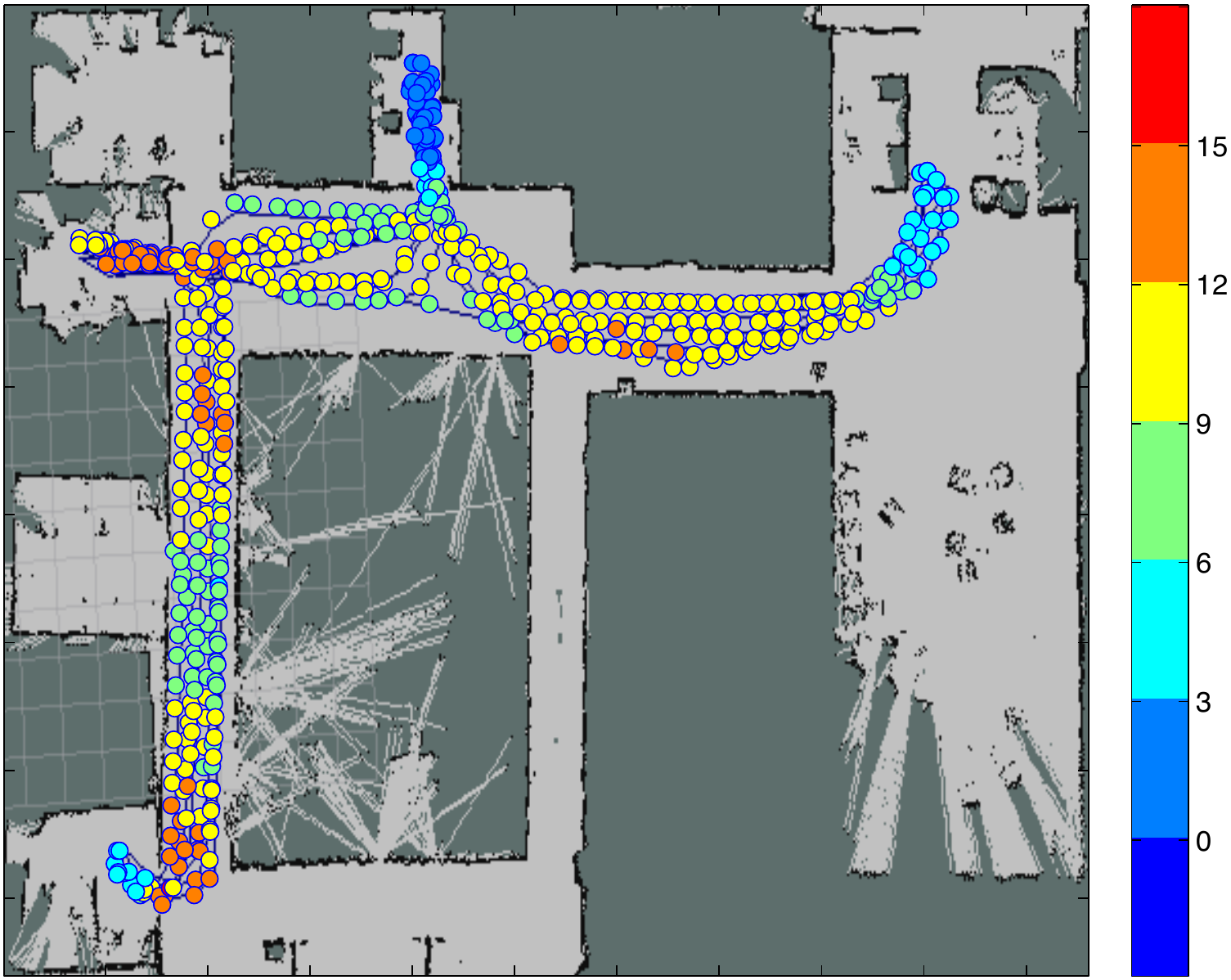}
\vspace{-1mm}\\
\hspace{-2mm}{(a)} & \hspace{-3mm}{(b)} 
\vspace{-3mm}
\end{tabular}
\caption{(a) Pioneer $3$-DX mobile robot trajectory of about $280$~m in the SMART
FM IRG office/lab generated by AMCL package in ROS, along which (b) $561$ relative light ($\%$) observations/data are gathered at locations denoted by small colored circles.}\vspace{-4mm}
\label{p3dxtraj}
\end{figure}
The \emph{adaptive Monte Carlo localization} (AMCL) package in the \emph{Robot Operating System} (ROS) is run on a Pioneer $3$-DX mobile robot mounted with a SICK LMS200 laser rangefinder to determine its trajectory (Fig.~\ref{p3dxtraj}a) and the $561$ locations at which the relative light measurements are taken (Fig.~\ref{p3dxtraj}b); these locations are assumed to be the ground truth.
GP-Localize, SoD-Truncate, and SoD-Even achieve, respectively, localization errors of $2.1$~m, $5.4$~m, and $4.6$~m averaged over all $561$ time steps and $3$ runs.
%
%
%


Fig.~\ref{fig:time}b shows the time incurred by GP-Localize, SoD-Truncate, SoD-Even, full GP, and offline PITC at each time step using $100$ particles averaged over $5$ runs.
It can be seen that, with more time steps, the time incurred by 
full GP, offline PITC, and SoD-Even increase while that of GP-Localize and SoD-Truncate remain constant.
GP-Localize is clearly much more scalable (i.e., constant time) in $t$ than full GP and offline PITC. Though it incurs slightly more time than SoD-Truncate and SoD-Even, it can localize significantly better (Sections~\ref{wssf} and~\ref{ieq}). 
%
\section{Conclusion} 
\label{sec:conclusion}
This paper describes the GP-Localize algorithm for persistent robot localization whose observation model is represented by our novel online sparse GP, thus achieving constant time and memory (i.e., independent of the size of the data) per filtering step.
We theoretically analyze the equivalence of our online sparse GP to the online learning variant of offline PITC. We empirically demonstrate that GP-Localize outperforms existing GP localization algorithms in terms of localization performance and scalability 
and achieves robustness by exploiting multiple fields.
Besides using our online sparse GP for persistent robot localization, note that it can in fact be applied to a wide variety of applications and are especially desirable (i.e., due to runtime and memory being independent of the size of data) for tasks with data streaming in over time or real-time requirements. Some robotic tasks include adaptive sampling, information gathering, learning of robot arm control \cite{LowAAMAS02,LowICRA02,LowNECO05}, visual tracking and head pose estimation for human-robot interaction. For non-robotic applications, they include traffic and weather prediction, online action recognition, online recommendation systems, online classification, among others.

A limitation of GP-Localize, as observed in our experiments, is that it does not localize well in near-constant fields, which is expected. So, in our future work, we plan to generalize our algorithm to handle richer, high-dimensional sensing data like laser scans and camera images \cite{Natarajan_2012,Natarajan_2012_ext,Natarajan_2014}.
We also like to investigate the effect of varying the slice size $\tau$ on the localization error of GP-Localize empirically
%
and remove the assumption of independence between fields by exploiting techniques like multi-output GPs and co-kriging for modeling their correlation.
Lastly, as mentioned in Section~\ref{sect:expt}, the hyperparameters of each GP are learned using the data by maximizing the log marginal likelihood. The sparse approximation method employed by offline PITC to improve the scalability of the full GP can be similarly applied to computing such a log marginal likelihood scalably, as explained in \cite{joaquin2005} (i.e., equation $30$ in Section $9$). Since our online sparse GP is the online variant of the offline PITC, the log marginal likelihood can be computed and maximized in an online manner as well. The exact details will be specified in the future extension of this work.
\section*{Acknowledgments}
This work was supported by Singapore-MIT Alliance for Research and Technology Subaward Agreement No. $41$ R-$252$-$000$-$527$-$592$.

\bibliographystyle{aaai}
\bibliography{references}

\begin{thebibliography}{}

\bibitem[\protect\citeauthoryear{Bodik \bgroup et al\mbox.\egroup
  }{2004}]{Guestrin04}
Bodik, P.; Guestrin, C.; Hong, W.; Madden, S.; Paskin, M.; and Thibaux, R.
\newblock 2004.
\newblock http://www.select.cs.cmu.edu/data/labapp3/index.html.

\bibitem[\protect\citeauthoryear{Brooks, Makarenko, and
  Upcroft}{2008}]{Brooks2008}
Brooks, A.; Makarenko, A.; and Upcroft, B.
\newblock 2008.
\newblock {G}aussian process models for indoor and outdoor sensor-centric robot
  localization.
\newblock {\em {IEEE} Trans. Robotics} 24(6):1341--1351.

\bibitem[\protect\citeauthoryear{Cao, Low, and Dolan}{2013}]{LowAAMAS13}
Cao, N.; Low, K.~H.; and Dolan, J.~M.
\newblock 2013.
\newblock Multi-robot informative path planning for active sensing of
  environmental phenomena: A tale of two algorithms.
\newblock In {\em Proc. {AAMAS}},  7--14.

\bibitem[\protect\citeauthoryear{Chen and Guestrin}{2007}]{Guestrin07}
Chen, K., and Guestrin, C.
\newblock 2007.
\newblock http://www.select.cs.cmu.edu/data/index.html.

\bibitem[\protect\citeauthoryear{Chen \bgroup et al\mbox.\egroup
  }{2012}]{LowUAI12}
Chen, J.; Low, K.~H.; Tan, C. K.-Y.; Oran, A.; Jaillet, P.; Dolan, J.~M.; and
  Sukhatme, G.~S.
\newblock 2012.
\newblock Decentralized data fusion and active sensing with mobile sensors for
  modeling and predicting spatiotemporal traffic phenomena.
\newblock In {\em Proc. UAI},  163--173.

\bibitem[\protect\citeauthoryear{Chen \bgroup et al\mbox.\egroup
  }{2013}]{LowUAI13}
Chen, J.; Cao, N.; Low, K.~H.; Ouyang, R.; Tan, C. K.-Y.; and Jaillet, P.
\newblock 2013.
\newblock Parallel {Gaussian} process regression with low-rank covariance
  matrix approximations.
\newblock In {\em Proc. UAI},  152--161.

\bibitem[\protect\citeauthoryear{Chen, Low, and Tan}{2013}]{LowRSS13}
Chen, J.; Low, K.~H.; and Tan, C. K.-Y.
\newblock 2013.
\newblock {Gaussian} process-based decentralized data fusion and active sensing
  for mobility-on-demand system.
\newblock In {\em Proc. {RSS}}.

\bibitem[\protect\citeauthoryear{Csat\'{o} and Opper}{2002}]{Csato02}
Csat\'{o}, L., and Opper, M.
\newblock 2002.
\newblock Sparse online {Gaussian} processes.
\newblock {\em Neural Computation} 14(2):641--669.

\bibitem[\protect\citeauthoryear{Dolan \bgroup et al\mbox.\egroup
  }{2009}]{LowSPIE09}
Dolan, J.~M.; Podnar, G.; Stancliff, S.; Low, K.~H.; Elfes, A.; Higinbotham,
  J.; Hosler, J.~C.; Moisan, T.~A.; and Moisan, J.
\newblock 2009.
\newblock Cooperative aquatic sensing using the telesupervised adaptive ocean
  sensor fleet.
\newblock In {\em Proc. {SPIE} Conference on Remote Sensing of the Ocean, Sea
  Ice, and Large Water Regions}, volume 7473.

\bibitem[\protect\citeauthoryear{Ferris, Fox, and Lawrence}{2007}]{Ferris2007}
Ferris, B.; Fox, D.; and Lawrence, N.
\newblock 2007.
\newblock {W}i{F}i-{SLAM} using {G}aussian process latent variable models.
\newblock In {\em Proc. {IJCAI}},  2480--2485.

\bibitem[\protect\citeauthoryear{Ferris, H\"{a}hnel, and
  Fox}{2006}]{Ferris2006}
Ferris, B.; H\"{a}hnel, D.; and Fox, D.
\newblock 2006.
\newblock {G}aussian processes for signal strength-based location estimation.
\newblock In {\em Proc. {RSS}}.

\bibitem[\protect\citeauthoryear{Gerkey, Vaughan, and
  Howard}{2003}]{Gerkey_2003}
Gerkey, B.~P.; Vaughan, R.~T.; and Howard, A.
\newblock 2003.
\newblock The {Player/Stage} project: Tools for multi-robot and distributed
  sensor systems.
\newblock In {\em Proc. ICAR},  317--323.

\bibitem[\protect\citeauthoryear{Hoang \bgroup et al\mbox.\egroup
  }{2014}]{LowICML14}
Hoang, T.~N.; Low, K.~H.; Jaillet, P.; and Kankanhalli, M.
\newblock 2014.
\newblock Nonmyopic $\epsilon$-{Bayes}-optimal active learning of {Gaussian}
  processes.
\newblock In {\em Proc. {ICML}}.

\bibitem[\protect\citeauthoryear{Ko and Fox}{2009a}]{Ko2008}
Ko, J., and Fox, D.
\newblock 2009a.
\newblock {GP}-{B}ayes{F}ilters: {B}ayesian filtering using {G}aussian process
  prediction and observation models.
\newblock {\em Autonomous Robots} 27(1):75--90.

\bibitem[\protect\citeauthoryear{Ko and Fox}{2009b}]{Ko2009}
Ko, J., and Fox, D.
\newblock 2009b.
\newblock Learning {GP}-{B}ayes{F}ilters via {G}aussian process latent variable
  models.
\newblock In {\em Proc. {RSS}}.

\bibitem[\protect\citeauthoryear{Krause, Singh, and Guestrin}{2008}]{Krause08}
Krause, A.; Singh, A.; and Guestrin, C.
\newblock 2008.
\newblock Near-optimal sensor placements in {Gaussian} processes: Theory,
  efficient algorithms and empirical studies.
\newblock {\em JMLR} 9:235--284.

\bibitem[\protect\citeauthoryear{Low \bgroup et al\mbox.\egroup
  }{2007}]{LowICRA07}
Low, K.~H.; Gordon, G.~J.; Dolan, J.~M.; and Khosla, P.
\newblock 2007.
\newblock Adaptive sampling for multi-robot wide-area exploration.
\newblock In {\em Proc. {IEEE ICRA}},  755--760.

\bibitem[\protect\citeauthoryear{Low \bgroup et al\mbox.\egroup
  }{2012}]{LowAAMAS12}
Low, K.~H.; Chen, J.; Dolan, J.~M.; Chien, S.; and Thompson, D.~R.
\newblock 2012.
\newblock Decentralized active robotic exploration and mapping for
  probabilistic field classification in environmental sensing.
\newblock In {\em Proc. {AAMAS}},  105--112.

\bibitem[\protect\citeauthoryear{Low, Dolan, and Khosla}{2008}]{LowAAMAS08}
Low, K.~H.; Dolan, J.~M.; and Khosla, P.
\newblock 2008.
\newblock Adaptive multi-robot wide-area exploration and mapping.
\newblock In {\em Proc. {AAMAS}},  23--30.

\bibitem[\protect\citeauthoryear{Low, Dolan, and Khosla}{2009}]{LowICAPS09}
Low, K.~H.; Dolan, J.~M.; and Khosla, P.
\newblock 2009.
\newblock Information-theoretic approach to efficient adaptive path planning
  for mobile robotic environmental sensing.
\newblock In {\em Proc. {ICAPS}},  233--240.

\bibitem[\protect\citeauthoryear{Low, Dolan, and Khosla}{2011}]{LowAAMAS11}
Low, K.~H.; Dolan, J.~M.; and Khosla, P.
\newblock 2011.
\newblock Active {Markov} information-theoretic path planning for robotic
  environmental sensing.
\newblock In {\em Proc. {AAMAS}},  753--760.

\bibitem[\protect\citeauthoryear{Low, Leow, and {Ang, Jr.}}{2002a}]{LowAAMAS02}
Low, K.~H.; Leow, W.~K.; and {Ang, Jr.}, M.~H.
\newblock 2002a.
\newblock A hybrid mobile robot architecture with integrated planning and
  control.
\newblock In {\em Proc. {AAMAS}},  219--226.

\bibitem[\protect\citeauthoryear{Low, Leow, and {Ang, Jr.}}{2002b}]{LowICRA02}
Low, K.~H.; Leow, W.~K.; and {Ang, Jr.}, M.~H.
\newblock 2002b.
\newblock Integrated planning and control of mobile robot with self-organizing
  neural network.
\newblock In {\em Proc. {IEEE ICRA}},  3870--3875.

\bibitem[\protect\citeauthoryear{Low, Leow, and {Ang, Jr.}}{2005}]{LowNECO05}
Low, K.~H.; Leow, W.~K.; and {Ang, Jr.}, M.~H.
\newblock 2005.
\newblock An ensemble of cooperative extended {Kohonen} maps for complex robot
  motion tasks.
\newblock {\em Neural Comput.} 17(6):1411--1445.

\bibitem[\protect\citeauthoryear{Natarajan \bgroup et al\mbox.\egroup
  }{2012a}]{Natarajan_2012}
Natarajan, P.; Hoang, T.~N.; Low, K.~H.; and Kankanhalli, M.
\newblock 2012a.
\newblock Decision-theoretic approach to maximizing observation of multiple
  targets in multi-camera surveillance.
\newblock In {\em Proc. AAMAS},  155--162.

\bibitem[\protect\citeauthoryear{Natarajan \bgroup et al\mbox.\egroup
  }{2012b}]{Natarajan_2012_ext}
Natarajan, P.; Hoang, T.~N.; Low, K.~H.; and Kankanhalli, M.
\newblock 2012b.
\newblock Decision-theoretic coordination and control for active multi-camera
  surveillance in uncertain, partially observable environments.
\newblock In {\em Proc. {ICDSC}}.

\bibitem[\protect\citeauthoryear{Natarajan, Low, and
  Kankanhalli}{2014}]{Natarajan_2014}
Natarajan, P.; Low, K.~H.; and Kankanhalli, M.
\newblock 2014.
\newblock Decision-theoretic approach to maximizing fairness in multi-target
  observation in multi-camera surveillance.
\newblock In {\em Proc. AAMAS}.

\bibitem[\protect\citeauthoryear{Ouyang \bgroup et al\mbox.\egroup
  }{2014}]{LowAAMAS14}
Ouyang, R.; Low, K.~H.; Chen, J.; and Jaillet, P.
\newblock 2014.
\newblock Multi-robot active sensing of non-stationary {Gaussian} process-based
  environmental phenomena.
\newblock In {\em Proc. {AAMAS}}.

\bibitem[\protect\citeauthoryear{Podnar \bgroup et al\mbox.\egroup
  }{2010}]{LowAeroconf10}
Podnar, G.; Dolan, J.~M.; Low, K.~H.; and Elfes, A.
\newblock 2010.
\newblock Telesupervised remote surface water quality sensing.
\newblock In {\em Proc. {IEEE} Aerospace Conference}.

\bibitem[\protect\citeauthoryear{{Qui\~{n}onero-Candela} and
  Rasmussen}{2005}]{joaquin2005}
{Qui\~{n}onero-Candela}, J., and Rasmussen, C.~E.
\newblock 2005.
\newblock A unifying view of sparse approximate {G}aussian process regression.
\newblock {\em JMLR} 6:1939--1959.

\bibitem[\protect\citeauthoryear{Rasmussen and Williams}{2006}]{Rasmussen2006}
Rasmussen, C.~E., and Williams, C. K.~I.
\newblock 2006.
\newblock {\em Gaussian Processes for Machine Learning}.
\newblock Cambridge, MA: MIT Press.

\bibitem[\protect\citeauthoryear{Snelson}{2007}]{SnelsonThesis07}
Snelson, E.~L.
\newblock 2007.
\newblock {\em Flexible and efficient {Gaussian} process models for machine
  learning}.
\newblock {Ph.D. Thesis}, University College London, London, UK.

\bibitem[\protect\citeauthoryear{Thrun, Burgard, and Fox}{2005}]{Thrun2005}
Thrun, S.; Burgard, W.; and Fox, D.
\newblock 2005.
\newblock {\em Probabilistic Robotics}.
\newblock Cambridge, MA: MIT Press.

\bibitem[\protect\citeauthoryear{Yu \bgroup et al\mbox.\egroup
  }{2012}]{LowIAT12}
Yu, J.; Low, K.~H.; Oran, A.; and Jaillet, P.
\newblock 2012.
\newblock Hierarchical {Bayesian} nonparametric approach to modeling and
  learning the wisdom of crowds of urban traffic route planning agents.
\newblock In {\em Proc. {IAT}},  478--485.

\end{thebibliography}

\if\myproof1
\clearpage
\appendix
\section{Derivation of Equation~\ref{fgp}}
\label{derivation}
$$
\hspace{-1.8mm}
\begin{array}{l}
\displaystyle p(z_{t} | x_t, u_{1:t}, z_{{1:t-1}}) \\
\displaystyle = \int p(x_{0:t-1}, z_{t}| x_t, u_{1:t}, z_{{1:t-1}})\ \text{d} x_{0:t-1}\\
\displaystyle = \int p(x_{0:t-1}| x_t, u_{1:t}, z_{{1:t-1}})\ p(z_{t}|x_t, x_{1:t-1}, z_{{1:t-1}})\ \text{d} x_{0:t-1}\\
\displaystyle = \int \prod_{i=1}^t p(x_{i-1}| x_i, u_{1:t}, z_{{1:t-1}})\ p(z_{t}|x_t, x_{1:t-1}, z_{{1:t-1}})\ \text{d} x_{0:t-1}\\
\displaystyle = \int \prod_{i=1}^t \frac{p(x_{i-1}| u_{1:t}, z_{{1:t-1}})\ p(x_i|u_i,x_{i-1})}{p(x_i|u_{1:t},z_{{1:t-1}})}\\
\quad\quad p(z_{t}|x_t, x_{1:t-1}, z_{{1:t-1}})\ \text{d} x_{0:t-1}\\
\displaystyle = \eta\int p(x_0)\prod_{i=1}^t p(x_i|u_i, x_{i-1})\ p(z_{t}|x_t, x_{1:t-1}, z_{{1:t-1}})\ \text{d} x_{0:t-1}
\end{array} 
$$
where $\eta = 1/p(x_t|u_{1:t},z_{{1:t-1}})$.
The third equality follows from the chain rule for probability followed by an assumption of conditional independence:
$p(x_{i-1}| x_{i:t}, u_{1:t}, z_{{1:t-1}})=p(x_{i-1}| x_i, u_{1:t}, z_{{1:t-1}})$ for $i=1,\ldots,t$. The fourth equality is due to the Bayes rule.
\section{Proof of Theorem~\ref{woohoo}}
\label{derivation2}
Since $\set{D}_n = x_{(n-1)\tau+1:n\tau}^{c}$ (Definition~\ref{def:ls}) and $t=N\tau +1$, $\set{D} = \bigcup_{n=1}^N\set{D}_n = \bigcup_{n=1}^N x_{(n-1)\tau+1:n\tau}^{c} = x_{1:N\tau}^{c} = x_{1:t-1}^{c}$.
Let us first simplify the $\Gamma_{x_t\set{D}}\left(\Gamma_{\set{D}\set{D}}+\Lambda\right)^{-1}$ term on the right-hand side expressions of $\mu_{x_t|\set{D}}^{\text{PITC}}=\mu^{\text{PITC}}_{x_t|x_{1:t-1}^{c}}$ \eqref{PITCmean} and $\sigma_{x_t x_t|\set{D}}^{\text{PITC}} =\sigma^{\text{PITC}}_{x_t x_t| x^c_{1:t-1}}$ \eqref{PITCvar}.
  \begin{equation} 
    \begin{array}{l} 
\left(\Gamma_{\set{D}\set{D}}+\Lambda\right)^{-1}\\
\displaystyle = \left(\Sigma_{\set{D}\set{S}}\Sigma_{\set{S}\set{S}}^{-1}\Sigma_{\set{S}\set{D}}+\Lambda\right)^{-1}\\
\displaystyle = \Lambda^{-1}-\Lambda^{-1}\Sigma_{\set{D}\set{S}}
    \left(\Sigma_{\set{S}\set{S}}+\Sigma_{\set{S}\set{D}}\Lambda^{-1}\Sigma_{\set{D}\set{S}}\right)^{-1}
    \Sigma_{\set{S}\set{D}}\Lambda^{-1}\\ 
\displaystyle =\Lambda^{-1}-\Lambda^{-1}\Sigma_{\set{D}\set{S}}\left({\Sigma}^N_{\circled{a}}\right)^{-1}\Sigma_{\set{S}\set{D}}\Lambda^{-1} \ .
  \end{array} 
    \label{eq:pitck} 
\end{equation}
The second equality is due to the matrix inversion lemma.
The last equality  follows from
\begin{equation}
    \begin{array}{l}
\displaystyle\Sigma_{\set{S}\set{S}}+\Sigma_{\set{S}\set{D}}\Lambda^{-1}\Sigma_{\set{D}\set{S}}\\
=\displaystyle \Sigma_{\set{S}\set{S}}+\sum_{n=1}^{N}\Sigma_{\set{S}\set{D}_n}\Sigma_{\set{D}_n \set{D}_n|\set{S}}^{-1}\Sigma_{\set{D}_n\set{S}}\\
= \displaystyle \Sigma_{\set{S}\set{S}}+\sum_{n=1}^{N}{\Sigma}^n_{\circled{s}} =\displaystyle {\Sigma}^N_{\circled{a}} \ .
  \end{array} 
    \label{eq:aa} 
\end{equation}
From \eqref{eq:pitck},
  \begin{equation} 
    \begin{array}{l} 
\Gamma_{x_t\set{D}}\left(\Gamma_{\set{D}\set{D}}+\Lambda\right)^{-1}\\
=\displaystyle\Sigma_{x_t\set{S}}\Sigma_{\set{S}\set{S}}^{-1}\Sigma_{\set{S}\set{D}}\left(\Lambda^{-1}-\Lambda^{-1}\Sigma_{\set{D}\set{S}}\left({\Sigma}^N_{\circled{a}}\right)^{-1}\Sigma_{\set{S}\set{D}}\Lambda^{-1}\right)\\
=\displaystyle\Sigma_{x_t\set{S}}\Sigma_{\set{S}\set{S}}^{-1}\left({\Sigma}^N_{\circled{a}}-\Sigma_{\set{S}\set{D}}\Lambda^{-1}\Sigma_{\set{D}\set{S}}\right)\left({\Sigma}^N_{\circled{a}}\right)^{-1}\Sigma_{\set{S}\set{D}}\Lambda^{-1}\vspace{1mm}\\
=\displaystyle\Sigma_{x_t\set{S}}\left({\Sigma}^N_{\circled{a}}\right)^{-1}\Sigma_{\set{S}\set{D}}\Lambda^{-1}\ .
  \end{array} 
    \label{eq:pitck2} 
\end{equation}
The third equality is due to (\ref{eq:aa}).

From \eqref{PITCmean},
%
\begin{equation*}
  \begin{array}{rl} 
\mu^{\text{PITC}}_{x_t|x_{1:t-1}^{c}}
\hspace{-2.8mm} &= \mu^{\text{PITC}}_{x_t| \set{D}}\\
\hspace{-2.8mm} &=\displaystyle\mu_{x_t}+\Gamma_{x_t\set{D}}\left(\Gamma_{\set{D}\set{D}}+\Lambda\right)^{-1}\left(z_\set{D}-\mu_\set{D} \right)\\
\hspace{-2.8mm} &=\displaystyle \mu_{x_t}+\Sigma_{x_t\set{S}}\left({\Sigma}^N_{\circled{a}}\right)^{-1}\Sigma_{\set{S}\set{D}}\Lambda^{-1}\left(z_\set{D}-\mu_\set{D} \right)\\
\hspace{-2.8mm} &=\displaystyle \mu_{x_t}+\Sigma_{x_t\set{S}}\left({\Sigma}^N_{\circled{a}}\right)^{-1}{\mu}^N_{\circled{a}}\\
\hspace{-2.8mm} &= \widetilde{\mu}_{x_t} \ .
  \end{array} 
  \label{eq:pitc.mur}
\end{equation*}
The third equality is due to (\ref{eq:pitck2}).
The fourth equality follows from 
$
\Sigma_{\set{S}\set{D}}\Lambda^{-1}\left(z_\set{D}-\mu_\set{D} \right)
= \sum_{n=1}^{N}\Sigma_{\set{S}\set{D}_n}\Sigma_{\set{D}_n \set{D}_n|\set{S}}^{-1}\left(z_{\set{D}_n}-\mu_{\set{D}_n} \right) = \sum_{n=1}^{N} {\mu}^n_{\circled{s}} ={\mu}^N_{\circled{a}}$.

From \eqref{PITCvar},
%
\begin{equation*}
\hspace{-1.8mm}
  \begin{array}{rl} 
  \sigma^{\text{PITC}}_{x_t x_t| x^c_{1:t-1}} \hspace{-2.8mm} &=\sigma^{\text{PITC}}_{x_t x_t|\set{D}}\\
  \hspace{-2.8mm} & \displaystyle=\sigma_{x_t x_t}-\Gamma_{x_t\set{D}}\left(\Gamma_{\set{D}\set{D}}+\Lambda\right)^{-1}\Gamma_{\set{D}{x_t}}\\
\hspace{-2.8mm} & \displaystyle    =\sigma_{x_t x_t}-\Sigma_{x_t\set{S}}\left({\Sigma}^N_{\circled{a}}\right)^{-1}\Sigma_{\set{S}\set{D}}\Lambda^{-1}\Sigma_{\set{D}\set{S}}\Sigma_{\set{S}\set{S}}^{-1}\Sigma_{\set{S}{x_t}}\\
\hspace{-2.8mm} &  \displaystyle    =\sigma_{x_t x_t}-\Sigma_{x_t\set{S}}\left({\Sigma}^N_{\circled{a}}\right)^{-1}\left({\Sigma}^N_{\circled{a}}- \Sigma_{\set{S}\set{S}}\right)
  \Sigma_{\set{S}\set{S}}^{-1}\Sigma_{\set{S}{x_t}}\\
 \hspace{-2.8mm} &   \displaystyle =\sigma_{x_t x_t}-\Sigma_{x_t \set{S}}\left(\Sigma_{\set{S}\set{S}}^{-1}-\left({\Sigma}^N_{\circled{a}}\right)^{-1}\right)\Sigma_{\set{S} x_t}\\
 \hspace{-2.8mm} &   = \widetilde{\sigma}_{x_t x_t} \ .
  \end{array} 
      \label{eq:pitc.varr} 
\end{equation*}
The third and fourth equalities follow from \eqref{eq:pitck2} and \eqref{eq:aa}, respectively.
\section{Incremental Update Formulas of Gaussian Posterior Mean and Variance}
\label{sect:derivation3}
Using the matrix inversion lemma, the following incremental update formulas of the Gaussian posterior mean and variance can be obtained:
$$
\displaystyle\mu_{x|\set{D}\cup\set{D}'}\triangleq \mu_{x|\set{D}} + \Sigma_{x\set{D}'|\set{D}}\Sigma_{\set{D}'\set{D}'|\set{D}}^{-1}\left( z_{\set{D}'} - \mu_{\set{D}'|\set{D}} \right)
$$
$$
\displaystyle\sigma_{xx|\set{D}\cup\set{D}'}\triangleq\sigma_{xx|\set{D}}-\Sigma_{x\set{D}'|\set{D}}\Sigma_{\set{D}'\set{D}'|\set{D}}^{-1}\Sigma_{\set{D}'x|\set{D}}
$$
for all $\set{D},\set{D}'\subset\set{X}$ such that $\set{D}\cap\set{D}'=\emptyset$ and $x\in\set{X}\setminus(\set{D}\cup\set{D}')$.
\section{Proof Sketch of Theorem~\ref{timespace}}
\label{sect:complexity}
%
Firstly, $({\Sigma}^N_{\circled{a}})^{-1}$ in \eqref{pmeanold} and \eqref{pvarold} has to be evaluated at time steps $t=N\tau+1$ for $N\in\mathbb{Z}^{+}$. To avoid incurring $\set{O}(|\set{S}|^3)$ time to invert ${\Sigma}^N_{\circled{a}}$, the matrix inversion lemma can be used to obtain $({\Sigma}^N_{\circled{a}})^{-1}$ from $({\Sigma}^{N-1}_{\circled{a}})^{-1}$ (i.e., previously derived at time step $(N-1)\tau+1$) in $\set{O}(\tau |\set{S}|^2)$ time (i.e., assuming $\tau < |\set{S}|$) and $\set{O}(|\set{S}|^2)$ memory, as observed in the following derivation:
$$
\hspace{-1.8mm}
\begin{array}{l}
\left({\Sigma}^N_{\circled{a}}\right)^{-1} \\
= \left({\Sigma}^{N-1}_{\circled{a}} + \Sigma_{\set{S}\set{D}_N}\Sigma_{\set{D}_N \set{D}_N|\set{S}}^{-1}\Sigma_{\set{D}_N \set{S}}\right)^{-1}\\
= \left({\Sigma}^{N-1}_{\circled{a}}\right)^{-1} + \left({\Sigma}^{N-1}_{\circled{a}}\right)^{-1}
\Sigma_{\set{S}\set{D}_N} \\
\quad\left(\Sigma_{\set{D}_N \set{D}_N|\set{S}} \hspace{-0.5mm}+\hspace{-0.5mm} \Sigma_{\set{D}_N \set{S}} \left({\Sigma}^{N-1}_{\circled{a}}\right)^{\hspace{-0.5mm}-1}\hspace{-0.5mm} \Sigma_{\set{S}\set{D}_N} \right)^{\hspace{-1mm}-1}\hspace{-1mm}
\Sigma_{\set{D}_N \set{S}} \hspace{-0.5mm}\left({\Sigma}^{N-1}_{\circled{a}}\right)^{\hspace{-0.5mm}-1}\hspace{-1mm}.
\end{array}
$$
Since evaluating ${\mu}^N_{\circled{a}}$ in \eqref{pmeanold} also incurs $\set{O}(\tau |\set{S}|^2)$ time, $\Sigma^{-1}_{\set{S}\set{S}}$ can be evaluated \emph{prior} to exploration and localization while incurring $\set{O}(|\set{S}|^2)$ memory, and $\set{D}' =\emptyset$ at time steps $t=N\tau+1$, computing $\widetilde{\mu}_{x_t|\set{D}'} = \widetilde{\mu}_{x_t}$ \eqref{pmeanold} and $\widetilde{\sigma}_{x_t x_t|\set{D}'} = \widetilde{\sigma}_{x_t x_t}$ \eqref{pvarold} incur $\set{O}(\tau |\set{S}|^2)$ time and $\set{O}(|\set{S}|^2)$ memory at time steps $t=N\tau+1$ for $N\in\mathbb{Z}^{+}$. 

On the other hand, when $N\tau+1<t\leq(N+1)\tau$, $\widetilde{\Sigma}^{-1}_{\set{D}'\set{D}'}$ in \eqref{pmeannew} and \eqref{pvarnew} has to be evaluated.
Let $\set{D}'_{-}\triangleq x_{N\tau+1:t-2}^c$. Then, $\set{D}' = \set{D}'_{-}\cup x^c_{t-1}$.
To avoid incurring $\set{O}(|\set{D}'|^3)$ time to invert $\widetilde{\Sigma}_{\set{D}'\set{D}'}$, the matrix inversion lemma can again be used to obtain $\widetilde{\Sigma}^{-1}_{\set{D}'\set{D}'}$ from $\widetilde{\Sigma}^{-1}_{\set{D}'_-\set{D}'_-}$ (i.e., previously derived at time step $t-1$) in $\set{O}(|\set{S}|^2)$ time and $\set{O}(|\set{S}|^2)$ memory (i.e., $|\set{D}'|<\tau<|\set{S}|$), as observed in the following derivation:
$$
\hspace{-1.8mm}
\begin{array}{l}
\widetilde{\Sigma}^{-1}_{\set{D}'\set{D}'}\\
= \widetilde{\Sigma}^{-1}_{(\set{D}'_-\cup x^c_{t-1})(\set{D}'_-\cup x^c_{t-1})}\\
= \left( 
\begin{array}{cc}
\widetilde{\Sigma}_{\set{D}'_-\set{D}'_-} & \widetilde{\Sigma}_{\set{D}'_- x^c_{t-1}} \\
\widetilde{\Sigma}_{x^c_{t-1} \set{D}'_-} &  \widetilde{\sigma}_{x^c_{t-1} x^c_{t-1}}
\end{array}
\right)^{-1}\\
= \left( 
\begin{array}{cc}
\widetilde{\Sigma}^{-1}_{\set{D}'_-\set{D}'_-}+ \widetilde{\Sigma}^{-1}_{\set{D}'_-\set{D}'_-} \widetilde{\Sigma}_{\set{D}'_- x^c_{t-1}}\Psi & -\Psi^{\top} \\
-\Psi &  \widetilde{\sigma}_{x^c_{t-1} x^c_{t-1}|\set{D}'_-}
\end{array}
\right)
\end{array}
$$
where 
$$
\widetilde{\sigma}_{x^c_{t-1} x^c_{t-1}|\set{D}'_-} = \widetilde{\sigma}_{x^c_{t-1} x^c_{t-1}}-\widetilde{\Sigma}_{x^c_{t-1} \set{D}'_-}\widetilde{\Sigma}^{-1}_{\set{D}'_-\set{D}'_-}\widetilde{\Sigma}_{\set{D}'_- x^c_{t-1}}
$$
and
$\Psi=\widetilde{\sigma}_{x^c_{t-1} x^c_{t-1}|\set{D}'_-}\widetilde{\Sigma}_{x^c_{t-1} \set{D}'_-}\widetilde{\Sigma}^{-1}_{\set{D}'_-\set{D}'_-}$.
Note that computing $\widetilde{\Sigma}_{x^c_{t-1} \set{D}'_-}$ only incurs $\set{O}(|\set{S}|^2)$ time instead of $\set{O}(\set{D}'|\set{S}|^2)$ time because
$$
\hspace{-1.8mm}
\begin{array}{l}
\widetilde{\Sigma}_{x^c_{t-1} \set{D}'_-} \\
= {\Sigma}_{x^c_{t-1} \set{D}'_-} -
{\Sigma}_{x^c_{t-1} \set{S}}\left(\Sigma^{-1}_{\set{S}\set{S}} - \left({\Sigma}^N_{\circled{a}}\right)^{-1}\right)\Sigma_{\set{S} \set{D}'_-}\\
= {\Sigma}_{x^c_{t-1} \set{D}'_-} -
{\Sigma}_{x^c_{t-1} \set{S}}\\
\quad \left[ \left(\Sigma^{-1}_{\set{S}\set{S}}\hspace{-0.5mm} -\hspace{-0.5mm} \left({\Sigma}^N_{\circled{a}}\right)^{\hspace{-0.5mm}-1}\right)\hspace{-0.5mm}\Sigma_{\set{S} x_{N\tau+1:t-3}^c}, \left(\Sigma^{-1}_{\set{S}\set{S}}\hspace{-0.5mm} -\hspace{-0.5mm} \left({\Sigma}^N_{\circled{a}}\right)^{\hspace{-0.5mm}-1}\right)\hspace{-0.5mm}\Sigma_{\set{S} x_{t-2}^c}\right]
\end{array}
$$
and $(\Sigma^{-1}_{\set{S}\set{S}} - ({\Sigma}^N_{\circled{a}})^{-1})\Sigma_{\set{S} x_{N\tau+1:t-3}^c}$ is previously evaluated at time step $t-1$.
Similarly, evaluating $\widetilde{\mu}_{\set{D}'}$ in \eqref{pmeannew} only incurs $\set{O}(|\set{S}|^2)$ time instead of $\set{O}(\set{D}'|\set{S}|^2)$ time because
$\widetilde{\mu}_{\set{D}'} = ( \widetilde{\mu}^{\top}_{\set{D}'_-}, \widetilde{\mu}_{x^c_{t-1}})^{\top}$
and $\widetilde{\mu}_{\set{D}'_-}$ is previously evaluated at time step $t-1$.
Therefore, computing $\widetilde{\mu}_{x_t|\set{D}'}$ \eqref{pmeannew} and $\widetilde{\sigma}_{x_t x_t|\set{D}'}$ \eqref{pvarnew} incur only $\set{O}(|\set{S}|^2)$ time and $\set{O}(|\set{S}|^2)$ memory at time steps $t$ where $N\tau+1<t\leq(N+1)\tau$.
\pagebreak
\section{Additional figures for Section~\ref{wssf}}
\label{fig:wss1256}
\begin{figure}[h!]
\begin{tabular}{c}
\hspace{-2mm}\includegraphics[scale=0.351]{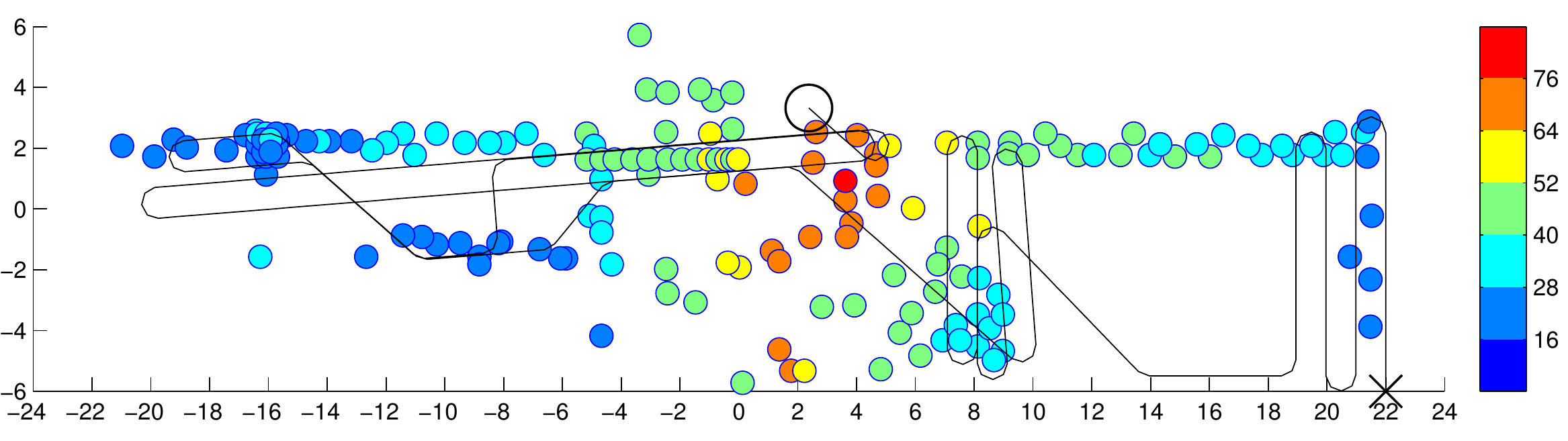}\vspace{-2mm}\\
\hspace{-2mm}{\scriptsize Access point $1$}\\
\hspace{-2mm}\includegraphics[scale=0.351]{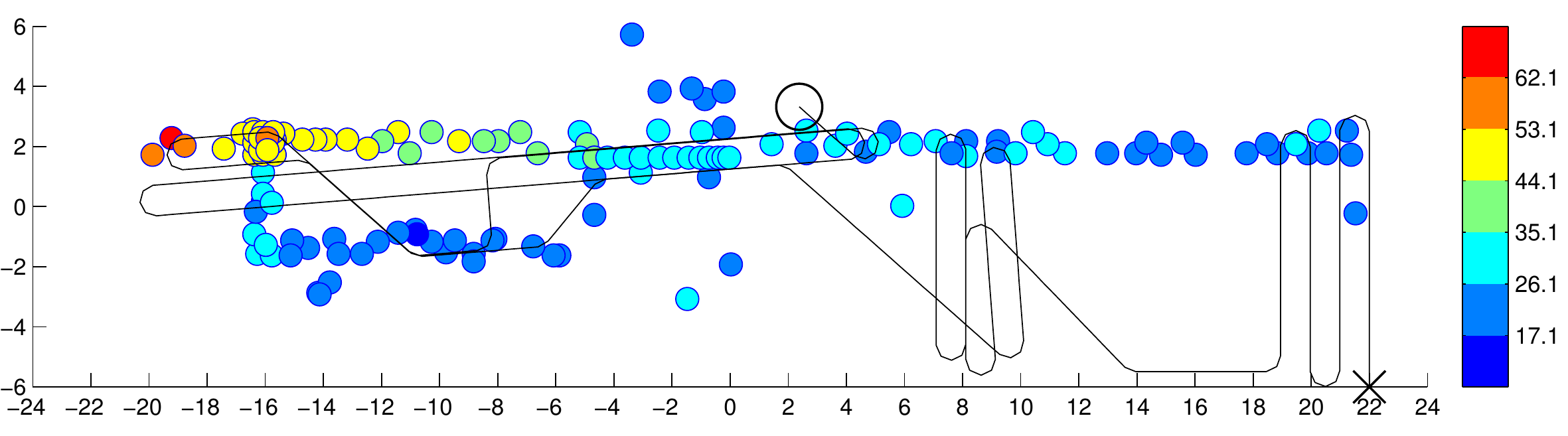}\vspace{-2mm}\\
\hspace{-2mm}{\scriptsize Access point $2$}\\
\hspace{-2mm}\includegraphics[scale=0.351]{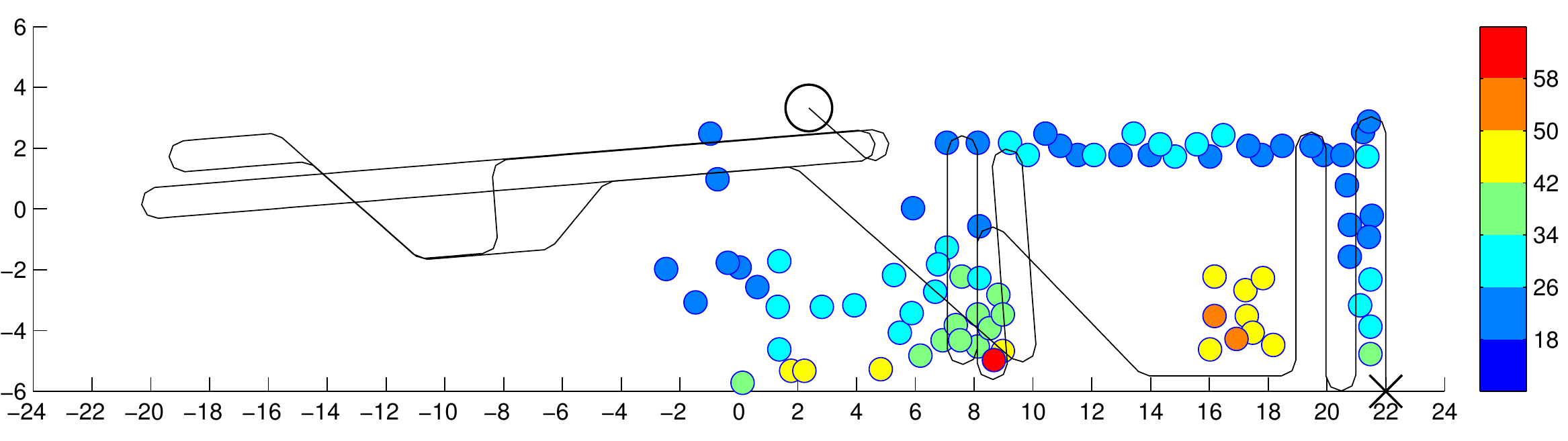}\vspace{-2mm}\\
\hspace{-2mm}{\scriptsize Access point $5$}\\
\hspace{-2mm}\includegraphics[scale=0.351]{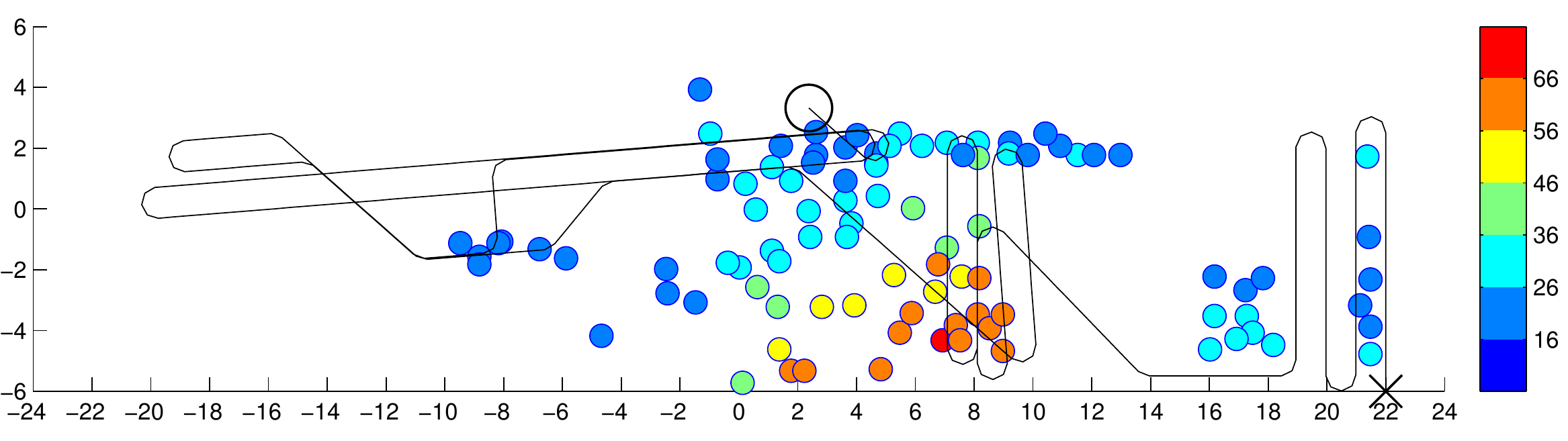}\vspace{-2mm}\\
\hspace{-2mm}{\scriptsize Access point $6$}\vspace{-4mm}
\end{tabular}
\caption{WSS data produced by WiFi APs $1$, $2$, $5$, and $6$ and measured at locations denoted by small colored circles with robot trajectory starting at `$\times$' and ending at `$\bigcirc$'.}\vspace{-5mm}
\label{fig:wifi-traj2}
\end{figure}
\fi
\end{document}